\documentclass[sigconf,natbib=false]{acmart}
\usepackage{algorithmic}
\usepackage{algorithm}
\usepackage{multicol}
\usepackage{multirow}
\usepackage{adjustbox}
\usepackage[table]{xcolor}

\AtBeginDocument{%
  }

\setcopyright{acmlicensed}

\copyrightyear{2026}
\acmYear{2026}
\acmDOI{XXXXXXX.XXXXXXX}
\acmConference[SAC'26]{The 41st ACM/SIGAPP Symposium on Applied Computing}{March 23--27, 2026}{Thessaloniki, Greece}
\acmISBN{979-X-XXXX-XXXX-X/26/03}

\RequirePackage[
  datamodel=acmdatamodel,
  style=acmnumeric,
  ]{biblatex}

\begin{document}

\title{Structured Extraction from Business Process Diagrams Using Vision-Language Models}
\renewcommand{\shorttitle}{Structured Extraction from Business Process Diagrams Using VLMs}

\author{Pritam Deka}
\email{p.deka@qub.ac.uk}
\orcid{0000-0002-8655-4762}
\affiliation{%
  \institution{Queen's University Belfast}
  \city{Belfast}
  \country{UK}
}

\author{Barry Devereux}
\email{B.Devereux@qub.ac.uk}
\orcid{0000-0003-2128-8632}
\affiliation{%
 \institution{Queen's University Belfast}
 \city{Belfast}
 \country{UK}}

\renewcommand{\shortauthors}{P. Deka and B. Devereux}

\begin{abstract}
 Business Process Model and Notation (BPMN) is a widely adopted standard for representing complex business workflows. While BPMN diagrams are often exchanged as visual images, existing methods primarily rely on XML representations for computational analysis. In this work, we present a pipeline that leverages Vision-Language Models (VLMs) to extract structured JSON representations of BPMN diagrams directly from images, without requiring source model files or textual annotations. We also incorporate optical character recognition (OCR) for textual enrichment and evaluate the generated element lists against ground truth data derived from the source XML files. Our approach enables robust component extraction in scenarios where original source files are unavailable. We benchmark multiple VLMs and observe performance improvements in several models when OCR is used for text enrichment. In addition, we conducted extensive statistical analyses of OCR-based enrichment methods and prompt ablation studies, providing a clearer understanding of their impact on model performance.

\end{abstract}

\ccsdesc[500]{Computing methodologies~Information extraction}

\keywords{Artificial Intelligence, Vision-Language Models, Diagram Understanding, Prompt Engineering, Structured Information Extraction}

\maketitle

\section{Introduction}
Business Process Model and Notation (BPMN) is a widely adopted standard for modeling and documenting business workflows \cite{allweyer2016bpmn}. 
It provides a visual formal language for processes through standardized visual elements-such as tasks (individual units of work), events (triggers or outcomes like start or end points), gateways (decision or branching logic), and sequence flows (arrows indicating control flow between elements)-that support both human interpretability and machine-readable semantics~\cite{kocbek2015business,allweyer2016bpmn,bpmn-standard}. BPMN diagrams are extensively used in enterprise environments, particularly for process documentation, compliance tracking, and training \cite{ko2009business}. However, these diagrams are often embedded in static formats such as PDF handbooks, training slides, or scanned documentation where the original XML-based BPMN source is not available. This lack of access to the structured source format presents a significant challenge for downstream applications that rely on semantic process data, such as simulation~\cite{laguna2018business}, automated migration~\cite{matos2009migrating}, conformance checking~\cite{van2012process}, or integration with workflow engines.

In this work, we present a vision-based approach for structured BPMN understanding using state-of-the-art vision-language models (VLMs)~\cite{meta2024llama3.2-vision,agrawal2024pixtral,hurst2024gpt,qwen2.5-VL,team2025gemma}. Our method treats BPMN diagrams as visual inputs and extracts structured JSON representations using a prompt that guides the model to detect and label each process element. The prompt instructs the model to identify and classify core components based on their visual layout. To improve coverage, especially in cases where labels are missing, low-contrast, or partially obscured, we introduce optional OCR-based enrichment using lightweight text extractors. This hybrid pipeline enables accurate recovery of BPMN semantics from images alone, without relying on the original XML source. 

\textbf{Our contributions are as follows:}
\begin{itemize}
  \item We propose a prompt-based pipeline for extracting structured representations from BPMN diagrams using VLM.
  \item We release a BPMN dataset with diagram–XML pairs for further research.
  \item We benchmark a range of VLMs and analyze OCR impact on extraction quality.
  \item We perform detailed ablation experiments and analysis to study the effect of different prompt types.
\end{itemize}

\section{Related Work}
Business process modeling, especially the construction and interpretation of BPMN diagrams, has attracted growing research interest with two main methodological approaches. 

\subsection{LLM-Driven BPMN Generation and Process Modeling}
Much of the recent progress in process automation comes from advances in LLMs. Early research relied on symbolic NLP and semantic role labeling to turn natural language into BPMN models~\cite{friedrich2011process}, but new systems use LLMs like GPT to improve accuracy and user interaction~\cite{sholiq2022generating,nour2024nala2bpmn,licardo2024method}. Tools such as Nala2BPMN~\cite{nour2024nala2bpmn} and~\cite{licardo2024method} combine these models with visualization toolkits, making process modeling more accessible.

Several modern frameworks take this further. ProMoAI~\cite{kourani2024promoai} lets users iteratively refine BPMN models from text and export them in multiple formats. The MAO Framework~\cite{lin2024mao} uses teams of LLM agents working in dialogue to collaboratively improve process models, while Opus~\cite{fagnoni2024opus} integrates LLMs with knowledge graphs to optimize business process outsourcing workflows. Although their goals differ, with ProMoAI focusing on user feedback, MAO on multi-agent teamwork, and Opus on domain knowledge, they all demonstrate how LLMs can automate and enhance BPMN generation.

At a finer level, studies like~\cite{bellan2022extracting} demonstrate that GPT-3 can extract process elements and relationships from text using prompt engineering alone, and~\cite{licardo2024method} shows that combining LLMs with NLP pre-processing tools like spaCy and BERT~\cite{devlin2019bert} improves extraction quality. These works highlight that, with the right prompts and minimal fine-tuning, LLMs can provide structured BPMN outputs without requiring specialized training.

Beyond just generating diagrams, LLMs have been applied to explanation and interactive analysis of process models. For example, \cite{minor2024retrieval} proposes combining case-based reasoning with LLMs for grounded explanations, and\cite{grohs2023large} and~\cite{kopke2024efficient} extend LLM use to tasks such as anomaly detection and process refinement. Some research takes a data-driven angle:~\cite{corradini2024technique} shows that process models can also be discovered from execution event logs, not just text, demonstrating the versatility of LLM-based approaches.

\subsection{VLMs for Diagram Understanding}
Alongside these text-driven methods, VLMs are rapidly advancing the ability to understand and extract information from visual process diagrams. VLMs have already shown success in visual question answering, image captioning, and layout understanding~\cite{hurst2024gpt,2023GPT4VisionSC,team2025gemma,fu2025vita}. Recently, researchers have started to apply these models to more structured visual reasoning, including tasks with symbolic diagrams~\cite{lu2023mathvista,zhang2024mathverse,chen2024far,roberts2024image2struct}.

The development of specialized datasets and benchmarks has been instrumental in advancing diagram understanding. In the domain of chart reasoning, resources such as ChartQA~\cite{masry2022chartqa}, ChartBench \cite{xu2023chartbench}, NovaChart \cite{hu2024novachart}, Chartx \cite{xia2024chartx} and ChartInstruct \cite{masry2024chartinstruct} have provided large-scale annotated data for training and evaluating models on quantitative and logical reasoning tasks. On top of these, models like ChartGPT~\cite{tian2024chartgpt}, Chartllama \cite{han2023chartllama}, Chartgemma \cite{masry2024chartgemma} and Chartvlm \cite{masry2024chartgemma} are adapted for chart-specific tasks. For process diagrams, benchmarks such as FlowchartQA~\cite{tannert2023flowchartqa} and FlowVQA \cite{singh2024flowvqa} provide diverse synthetic flowcharts paired with questions to evaluate the visual reasoning abilities of VLMs.
In contrast, BPMN-Redrawer~\cite{antinori2022bpmn} focuses on reconstructing executable BPMN XML representations directly from visual diagrams, a task that demands precise recovery of both entities and their relationships.

Building on this line of work, our research explores the more challenging and practical scenario of extracting structured BPMN representations directly from images, without relying on any accompanying text, underlying XML, or manual heuristics. We propose an approach that guides a VLM using a BPMN-specific prompt focused on visual and semantic characteristics, enabling generalization across various layouts and styles. This allows for schema-constrained recovery of all major BPMN elements (tasks, events, gateways, flows, pools, lanes) from static diagrams alone.

\section{Dataset Construction}
In this section, we describe how the dataset was constructed. Our goal was to create a clean, consistent, and visually aligned dataset that could serve as a strong benchmark for structured diagram understanding.

\subsection{Source Generation}
We built our dataset using BPMN 2.0 XML files collected from the publicly available \texttt{bpmn-for-research}\footnote{\url{https://github.com/camunda/bpmn-for-research}} repository \cite{bpmn-for-research}, which provides over 3,700 BPMN diagrams created during Camunda's training sessions. From this corpus, we selected only the English-language diagrams to maintain consistency in text content and labels. After filtering, we renamed all BPMN files using a standardized format (e.g., \texttt{process\_001.bpmn}) for consistent pairing with images. Each diagram was rendered to a high-resolution PNG using a custom script that leverages the \texttt{bpmn-js}\footnote{\url{https://github.com/bpmn-io/bpmn-js}} library from Camunda’s \texttt{bpmn.io} project~\cite{bpmn-js}. The final dataset consists of 202 diagram pairs, each containing a BPMN XML file and a cropped PNG image. All filenames are matched to allow straightforward alignment between formats. 

\subsection{Statistics and Format}

We split the 202 BPMN diagrams into training, development, and test sets using a \textbf{50:25:25 ratio}, resulting in \textbf{101 training}, \textbf{50 development}, and \textbf{51 test} diagrams. Unlike typical 80:20 or 70:30 splits, this choice was driven by our focus on \textbf{prompt-based evaluation} rather than model training. A larger test set allows for better analysis of model behavior across a diverse set of BPMN structures. Each sample includes a \textbf{BPMN diagram image (.png)} and its \textbf{corresponding XML file (.bpmn)}.

The dataset covers a diverse range of BPMN diagram types, including \textbf{single-pool processes}, \textbf{multi-lane and multi-pool layouts}, \textbf{tasks and subprocesses}, \textbf{gateways} (exclusive, inclusive, parallel), \textbf{events} (start, end, intermediate), and \textbf{sequence flows} with or without labels. This variety ensures the test set captures both simple and complex scenarios. The complete dataset with the splits, folder structures, and preprocessing scripts is available via \textbf{\url{https://github.com/pritamdeka/BPMN-VLM}}.

\section{Methodology}

The core methodology of our approach is centered around a prompt-based structured information extraction from BPMN diagrams using VLMs. The prompt includes a comprehensive list of extraction requirements and a sample JSON object to guide the model’s output structure, but does not provide any input–output example pairs. This approach ensures the VLM produces consistent, schema-compliant outputs across diverse BPMN diagram styles. We have used the same prompt across all the models.

\textbf{Problem Definition.}  
Let a BPMN diagram image be denoted as $I$. The objective is to extract a structured JSON representation of BPMN elements $E = \{ e_1, e_2, \dots, e_n \}$ where each element $e_i$ has a type, textual label, and attributes. We define the extraction as a mapping:
\begin{equation} \label{eq:1}
    f_\theta : (I, P, T) \mapsto R
\end{equation}

where $f_\theta$ is a VLM parameterized by $\theta$, $P$ is the input prompt, $T$ is optional OCR-enriched text, and $R$ is the raw textual response. 
\textbf{JSON Parsing.}  
The response $R$ is mapped to a structured representation $J(R)$ via a parsing operator:
\begin{equation} \label{eq:2}
    J(R) = 
\begin{cases}
\text{Valid JSON object} & \text{if parsing succeeds} \\
\varnothing & \text{otherwise}
\end{cases}
\end{equation}

We have explored two primary directions in our methodology which are explained in the subsequent subsections.

\subsection{VLM-Only Prompted Extraction}

In this setting, we input the BPMN diagram directly into a VLM accompanied by our carefully designed prompt that describes the visual symbols and semantics of BPMN elements (e.g., rectangles for tasks, diamonds for gateways). The model is expected to utilize its joint visual and textual understanding to process the image and output a structured JSON representation that mirrors the underlying process structure. This approach proves effective for simpler diagrams, where the layout is clean, components are spatially distinct, and most elements contain clearly visible text labels. Formally, the pipeline from Eq \ref{eq:1} and Eq \ref{eq:2} reduces to:
\begin{equation}
    R = f_\theta(I,P), \quad J(R) = \{e_1, e_2, \dots,e_n\}
\end{equation}

However, as diagram complexity increases, the model’s extraction accuracy declines noticeably. VLMs often struggle with: (1) \textbf{unlabeled or ambiguous elements}, especially gateways and events lacking names; (2) \textbf{dense layouts}, where overlapping elements in multi-pool or multi-lane diagrams cause confusion or omissions; (3) \textbf{low-contrast or edge-aligned text}, particularly on sequence flows; and (4) \textbf{resolution constraints}, as many VLMs rescale input images to fixed sizes, leading to the loss of fine-grained visual details in large or wide diagrams (see Section \ref{sec:error_dist} and Table \ref{tab:error_rates_overall} for a detailed empirical quantification of these errors).

\begin{figure}[h]
    \includegraphics[width=0.5\columnwidth]{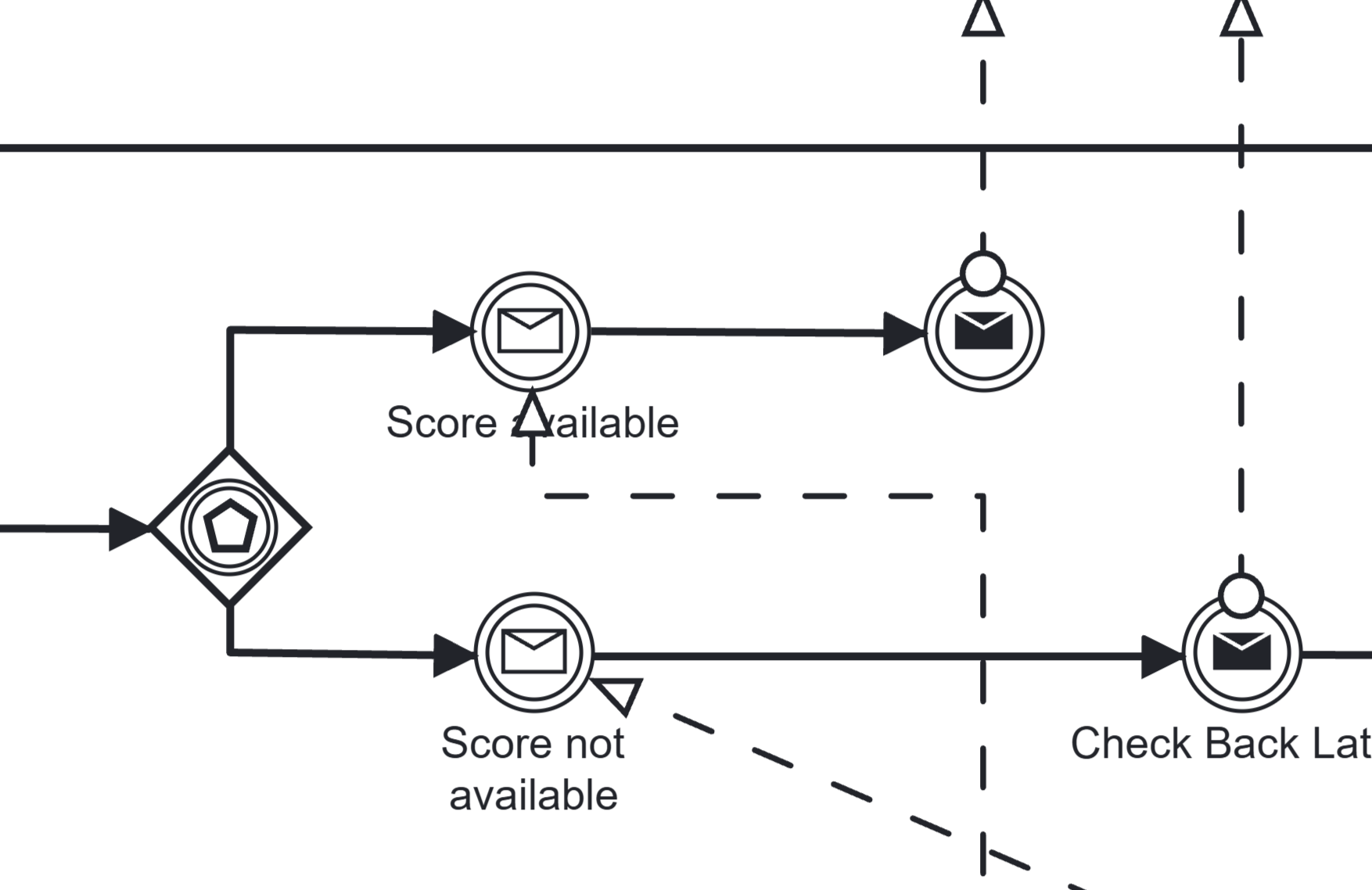}
    \caption{Cropped BPMN region with fine-grained cues}
    \label{fig:examplebpmn}
\end{figure}

Figure \ref{fig:examplebpmn} shows why VLM-only extraction may be error-prone. The model must identify: (1) the \textbf{pentagon} inside the gateway (an \textbf{EventBasedGateway}); (2) the difference between a solid \textbf{SequenceFlow} and a dashed \textbf{MessageFlow}; and (3) the envelope \textbf{IntermediateCatchEvent}. These small markers and line styles often vanish when the image is downsampled, which may lead to the misclassifications reported in Section~\ref{sec:error_dist}.

\subsection{OCR-Enriched Hybrid Extraction}

To address the limitations of the VLM-only extraction pipeline, we investigate whether OCR-based enrichment can serve as an effective fallback mechanism. Our hypothesis is that OCR may help recover textual labels that VLMs fail to detect, particularly for elements like sequence flows, gateways, or events that often lack clear or legible annotations. Our methodology comprises two stages: the VLM first generates a structured JSON from the BPMN diagram, while OCR tools such as Pix2Struct~\cite{lee2023pix2struct}, RapidOCR~\cite{RapidOCR2021}, or Tesseract~\cite{smith2007overview} extract text from the same image in parallel. During post-processing, an enrichment operation uses OCR outputs to fill missing component labels in the VLM’s output, improving entity completeness for visually ambiguous or illegible elements without altering the original structure. Formally, from Eq \ref{eq:1},
\begin{equation}
    R = f_\theta(I,P,T) \quad T = O_m(I)
\end{equation}
where $m \in \{\text{Pix2Struct}, \text{Tesseract}, \text{RapidOCR}\}$

Parsed results $J(R)$ are passed to an enrichment operator $E'$, which fills missing labels using OCR tokens:
\begin{equation}    
\hat{J} = E'(J(R),T).
\end{equation}
where, for any element $e \in J$ such that $\text{name}(e) = \emptyset$, we assign $\text{name}(e) \leftarrow \text{match}(T)$.

In both settings, we apply post-inference JSON validation and normalization to ensure consistency across models and support accurate alignment with ground truth XML during evaluation. This dual-branch architecture (VLM-only vs. VLM+OCR) allows us to analyze the trade-offs in model performance and identify cases where OCR integration leads to measurable improvements. Algorithm \ref{alg:bpmn} shows the pseudocode for our proposed method.


\begin{algorithm}[t]
\caption{Proposed BPMN Extraction Method}
\label{alg:bpmn}
\begin{algorithmic}[1]
\STATE \textbf{Input:} Image $I$, Prompt $P$, OCR flag $enable\_ocr$, OCR method $m$
\STATE \textbf{Output:} Structured JSON $\hat{J}$
\STATE Encode $I$ as base64 string $E(I)$
\IF{$enable\_ocr$}
    \STATE Extract text tokens $T \gets O_m(I)$
\ELSE
    \STATE $T \gets \emptyset$
\ENDIF
\STATE Generate raw response $R \gets f_\theta(I, P, T)$
\STATE Parse JSON $J \gets J(R)$
\IF{$J \neq \varnothing$}
    \STATE Enrich missing labels $\hat{J} \gets E'(J, T)$
\ELSE
    \STATE Save raw output $\hat{J} \gets R$
\ENDIF
\STATE \textbf{return} $\hat{J}$
\end{algorithmic}
\end{algorithm}

\subsection{Prompt Design}

We adopt a zero-shot, schema-constrained prompting strategy~\cite{roberts2024image2struct, medhi2024target, khan2024fine, scius2025zero}, providing only detailed instructions and an output example in the prompt, but no input–output pairs. The prompt was iteratively refined to maximize extraction accuracy and consistency. The prompt explicitly describes the visual characteristics and semantics of all BPMN components, specifies a strict output schema, and defines extraction rules for bounding boxes and label assignment. Without providing explicit examples, the prompt relies on detailed instructions and a template JSON structure to enforce consistent, machine-readable outputs across models. The full prompt is available in the project repository\footnote{\url{https://github.com/pritamdeka/BPMN-VLM/blob/main/prompts/prompt.txt}}.

\begin{table*}[h]
\caption{Strict evaluation results (names only) with precision, recall, and F1 across VLM-only and OCR-enriched pipelines \textbf{(best results in bold)}.}
\centering
\small
\setlength{\tabcolsep}{5pt}
\renewcommand{\arraystretch}{0.75}
\begin{adjustbox}{max width=\textwidth}
\begin{tabular}{lccc|ccc|ccc|ccc}
\toprule
\multirow{2}{*}{\textbf{Model}} &
\multicolumn{3}{c|}{\textbf{Only VLM}} &
\multicolumn{3}{c|}{\textbf{VLM + Pix2Struct}} &
\multicolumn{3}{c|}{\textbf{VLM + RapidOCR}} &
\multicolumn{3}{c}{\textbf{VLM + Tesseract}} \\
\cmidrule(lr){2-4} \cmidrule(lr){5-7} \cmidrule(lr){8-10} \cmidrule(lr){11-13}
& \textbf{P} & \textbf{R} & \textbf{F1} 
& \textbf{P} & \textbf{R} & \textbf{F1} 
& \textbf{P} & \textbf{R} & \textbf{F1} 
& \textbf{P} & \textbf{R} & \textbf{F1} \\
\midrule

Aya-vision-8B & 0.201 & 0.140 & 0.151 & 0.175 & 0.128 & 0.138 & 0.213 & 0.133 & 0.154 & 0.190 & 0.129 & 0.142 \\

Gemma-12B & 0.798 & 0.560 & 0.645 & 0.748 & 0.560 & 0.628 & 0.725 & 0.592 & 0.640 & 0.725 & 0.592 & 0.640 \\

Gemma-4B & 0.642 & 0.586 & 0.602 & 0.635 & 0.586 & 0.599 & 0.617 & 0.576 & 0.585 & 0.617 & 0.576 & 0.585 \\

GPT-4.1 & \textbf{0.829} & 0.794 & \textbf{0.807} 
& \textbf{0.783} & \textbf{0.810} & \textbf{0.792} 
& \textbf{0.781} & \textbf{0.802} & \textbf{0.787}
& \textbf{0.777} & 0.786 & 0.775 \\

GPT-4.1-mini & 0.823 & \textbf{0.819} & 0.817 
& 0.772 & 0.816 & 0.788 
& 0.766 & 0.814 & 0.784 
& 0.775 & \textbf{0.816} & \textbf{0.790} \\

GPT-4.1-nano & 0.712 & 0.535 & 0.598 & 0.661 & 0.529 & 0.576 & 0.681 & 0.522 & 0.579 & 0.645 & 0.504 & 0.553 \\

GPT-4o & 0.812 & 0.775 & 0.790 & 0.767 & 0.781 & 0.770 & 0.769 & 0.780 & 0.771 & 0.773 & 0.778 & 0.772 \\

GPT-4o-mini & 0.782 & 0.629 & 0.691 & 0.748 & 0.633 & 0.678 & 0.748 & 0.633 & 0.678 & 0.739 & 0.642 & 0.681 \\

LLaMA-3.2-11B & 0.373 & 0.334 & 0.344 & 0.375 & 0.302 & 0.322 & 0.394 & 0.327 & 0.343 & 0.344 & 0.309 & 0.318 \\

Mistral-small-3.1 & 0.830 & 0.781 & 0.802 & 0.776 & 0.778 & 0.772 & 0.773 & 0.788 & 0.776 & 0.787 & 0.774 & 0.774 \\

Pixtral-12B & 0.352 & 0.284 & 0.301 & 0.342 & 0.312 & 0.316 & 0.321 & 0.292 & 0.297 & 0.354 & 0.312 & 0.323 \\

Pixtral-large & 0.712 & 0.517 & 0.589 & 0.706 & 0.517 & 0.583 & 0.716 & 0.515 & 0.589 & 0.715 & 0.516 & 0.587 \\

Qwen-3B & 0.538 & 0.486 & 0.486 & 0.550 & 0.469 & 0.481 & 0.568 & 0.408 & 0.458 & 0.547 & 0.406 & 0.451 \\

Qwen-7B & 0.774 & 0.727 & 0.744 & 0.769 & 0.744 & 0.749 & 0.735 & 0.738 & 0.729 & 0.736 & 0.738 & 0.729 \\

\midrule
\multicolumn{13}{l}{\textbf{Non-VLM Baseline}} \\
BPMN Redrawer & 0.531 & 0.407 & 0.458 & -- & -- & -- & -- & -- & -- & -- & -- & -- \\
Sketch2Process & 0.787 & 0.630 & 0.693 & -- & -- & -- & -- & -- & -- & -- & -- & -- \\

\bottomrule
\end{tabular}
\end{adjustbox}
\label{tab:names-only-strict}
\end{table*}

\begin{table*}[h]
\caption{Strict vs. relaxed evaluation results (names + types) with precision, recall, and F1 across VLM-only and OCR-enriched pipelines \textbf{(best results in bold)}.}
\centering
\setlength{\tabcolsep}{3pt}
\renewcommand{\arraystretch}{0.90}
\begin{adjustbox}{max width=\textwidth}
\begin{tabular}{lcccccc|cccccc|cccccc|cccccc}
\toprule
\multirow{3}{*}{\textbf{Model}} &
\multicolumn{6}{c|}{\textbf{Only VLM}} &
\multicolumn{6}{c|}{\textbf{VLM + Pix2Struct}} &
\multicolumn{6}{c|}{\textbf{VLM + RapidOCR}} &
\multicolumn{6}{c}{\textbf{VLM + Tesseract}} \\
\cmidrule(lr){2-7} \cmidrule(lr){8-13} \cmidrule(lr){14-19} \cmidrule(lr){20-25}
& \multicolumn{3}{c}{Strict} & \multicolumn{3}{c|}{Relaxed}
& \multicolumn{3}{c}{Strict} & \multicolumn{3}{c|}{Relaxed}
& \multicolumn{3}{c}{Strict} & \multicolumn{3}{c|}{Relaxed}
& \multicolumn{3}{c}{Strict} & \multicolumn{3}{c}{Relaxed} \\
\cmidrule(lr){2-4} \cmidrule(lr){5-7}
\cmidrule(lr){8-10} \cmidrule(lr){11-13}
\cmidrule(lr){14-16} \cmidrule(lr){17-19}
\cmidrule(lr){20-22} \cmidrule(lr){23-25}
& \textbf{P} & \textbf{R} & \textbf{F1}
& \textbf{P} & \textbf{R} & \textbf{F1}
& \textbf{P} & \textbf{R} & \textbf{F1}
& \textbf{P} & \textbf{R} & \textbf{F1}
& \textbf{P} & \textbf{R} & \textbf{F1}
& \textbf{P} & \textbf{R} & \textbf{F1}
& \textbf{P} & \textbf{R} & \textbf{F1}
& \textbf{P} & \textbf{R} & \textbf{F1} \\
\midrule

Aya-vision-8B & 0.130 & 0.104 & 0.106 & 0.138 & 0.107 & 0.111 & 0.125 & 0.099 & 0.103 & 0.129 & 0.100 & 0.104 & 0.155 & 0.108 & 0.120 & 0.159 & 0.109 & 0.123 & 0.131 & 0.103 & 0.108 & 0.134 & 0.104 & 0.110 \\

Gemma-12B & 0.502 & 0.377 & 0.422 & 0.512 & 0.383 & 0.430 & 0.435 & 0.377 & 0.396 & 0.443 & 0.383 & 0.403 & 0.515 & 0.465 & 0.481 & 0.524 & 0.472 & 0.488 & 0.515 & 0.465 & 0.481 & 0.524 & 0.472 & 0.488 \\

Gemma-4B & 0.425 & 0.449 & 0.428 & 0.440 & 0.463 & 0.443 & 0.420 & 0.449 & 0.425 & 0.435 & 0.463 & 0.440 & 0.429 & 0.465 & 0.436 & 0.439 & 0.475 & 0.446 & 0.429 & 0.465 & 0.436 & 0.439 & 0.475 & 0.446 \\

GPT-4.1 & 0.718 & 0.702 & 0.707 & 0.733 & 0.716 & 0.721 & 0.612 & 0.709 & 0.652 & 0.629 & 0.727 & 0.670 & 0.604 & 0.694 & 0.642 & 0.624 & 0.716 & 0.663 & 0.580 & 0.674 & 0.619 & 0.593 & 0.688 & 0.633 \\

GPT-4.1-mini & 0.706 & \textbf{0.729} & \textbf{0.715} & 0.720 & \textbf{0.743} & \textbf{0.729} & 0.608 & \textbf{0.722} & 0.657 & 0.620 & \textbf{0.736} & 0.669 & 0.607 & \textbf{0.722} & \textbf{0.655} & 0.618 & \textbf{0.735} & 0.667 & 0.618 & 0.726 & 0.664 & 0.633 & 0.742 & \textbf{0.679} \\

GPT-4.1-nano & 0.638 & 0.492 & 0.544 & 0.647 & 0.498 & 0.551 & 0.556 & 0.482 & 0.506 & 0.562 & 0.488 & 0.511 & 0.541 & 0.458 & 0.486 & 0.552 & 0.468 & 0.497 & 0.537 & 0.456 & 0.481 & 0.545 & 0.463 & 0.489 \\

GPT-4o & 0.699 & 0.676 & 0.685 & 0.719 & 0.695 & 0.704 & 0.643 & 0.695 & \textbf{0.663} & 0.659 & 0.712 & \textbf{0.679} & 0.628 & 0.688 & 0.652 & 0.646 & 0.705 & \textbf{0.670} & \textbf{0.653} & 0.698 & \textbf{0.670} & \textbf{0.663} & 0.707 & \textbf{0.679} \\

GPT-4o-mini & 0.669 & 0.543 & 0.594 & 0.683 & 0.555 & 0.606 & \textbf{0.665} & 0.583 & 0.613 & \textbf{0.677} & 0.593 & 0.624 & \textbf{0.665} & 0.583 & 0.613 & \textbf{0.677} & 0.593 & 0.624 & 0.631 & 0.570 & 0.591 & 0.643 & 0.579 & 0.602 \\

LLaMA-3.2-11B & 0.280 & 0.259 & 0.263 & 0.282 & 0.261 & 0.265 & 0.303 & 0.257 & 0.269 & 0.304 & 0.258 & 0.270 & 0.282 & 0.250 & 0.256 & 0.286 & 0.253 & 0.259 & 0.275 & 0.260 & 0.261 & 0.275 & 0.260 & 0.261 \\

Mistral-small-3.1 & 0.696 & 0.690 & 0.691 & 0.712 & 0.705 & 0.706 & 0.604 & 0.669 & 0.631 & 0.624 & 0.689 & 0.651 & 0.613 & 0.689 & 0.645 & 0.633 & 0.710 & 0.666 & 0.614 & 0.681 & 0.642 & 0.635 & 0.702 & 0.662 \\

Pixtral-12B & 0.274 & 0.231 & 0.238 & 0.282 & 0.238 & 0.246 & 0.238 & 0.238 & 0.230 & 0.241 & 0.242 & 0.233 & 0.239 & 0.245 & 0.235 & 0.245 & 0.252 & 0.241 & 0.252 & 0.254 & 0.247 & 0.253 & 0.255 & 0.248 \\

Pixtral-large & 0.220 & 0.196 & 0.205 & 0.223 & 0.199 & 0.207 & 0.145 & 0.165 & 0.153 & 0.146 & 0.166 & 0.154 & 0.119 & 0.124 & 0.120 & 0.120 & 0.126 & 0.122 & 0.162 & 0.179 & 0.168 & 0.164 & 0.181 & 0.170 \\

Qwen-3B & 0.311 & 0.325 & 0.308 & 0.314 & 0.329 & 0.311 & 0.291 & 0.292 & 0.282 & 0.300 & 0.299 & 0.290 & 0.139 & 0.120 & 0.121 & 0.143 & 0.123 & 0.124 & 0.163 & 0.137 & 0.142 & 0.164 & 0.138 & 0.143 \\

Qwen-7B & 0.480 & 0.569 & 0.517 & 0.487 & 0.578 & 0.524 & 0.471 & 0.590 & 0.517 & 0.477 & 0.597 & 0.524 & 0.482 & 0.622 & 0.536 & 0.493 & 0.639 & 0.549 & 0.467 & 0.603 & 0.519 & 0.478 & 0.619 & 0.532 \\

\midrule
\multicolumn{25}{l}{\textbf{Non-VLM Baseline}} \\
BPMN Redrawer & 0.518 & 0.399 & 0.448 & 0.518 & 0.399 & 0.448 & -- & -- & -- & -- & -- & -- & -- & -- & -- & -- & -- & -- & -- & -- & -- & -- & -- & -- \\

Sketch2Process & \textbf{0.766} & 0.612 & 0.674 & \textbf{0.771} & 0.617 & 0.678 & -- & -- & -- & -- & -- & -- & -- & -- & -- & -- & -- & -- & -- & -- & -- & -- & -- & -- \\

\bottomrule
\end{tabular}
\end{adjustbox}
\label{tab:strict-relaxed-eval}
\end{table*}

\begin{table*}[h]
\caption{Strict vs. relaxed evaluation results (relations + types) with precision, recall, and F1 across VLM-only and OCR-enriched pipelines \textbf{(best results in bold)}.}
\centering
\setlength{\tabcolsep}{3pt}
\renewcommand{\arraystretch}{0.9}
\begin{adjustbox}{max width=\textwidth}
\begin{tabular}{lcccccc|cccccc|cccccc|cccccc}
\toprule
\multirow{3}{*}{\textbf{Model}} &
\multicolumn{6}{c|}{\textbf{Only VLM}} &
\multicolumn{6}{c|}{\textbf{VLM + Pix2Struct}} &
\multicolumn{6}{c|}{\textbf{VLM + RapidOCR}} &
\multicolumn{6}{c}{\textbf{VLM + Tesseract}} \\
\cmidrule(lr){2-7} \cmidrule(lr){8-13} \cmidrule(lr){14-19} \cmidrule(lr){20-25}
& \multicolumn{3}{c}{Strict} & \multicolumn{3}{c|}{Relaxed}
& \multicolumn{3}{c}{Strict} & \multicolumn{3}{c|}{Relaxed}
& \multicolumn{3}{c}{Strict} & \multicolumn{3}{c|}{Relaxed}
& \multicolumn{3}{c}{Strict} & \multicolumn{3}{c}{Relaxed} \\
\cmidrule(lr){2-4} \cmidrule(lr){5-7}
\cmidrule(lr){8-10} \cmidrule(lr){11-13}
\cmidrule(lr){14-16} \cmidrule(lr){17-19}
\cmidrule(lr){20-22} \cmidrule(lr){23-25}
& \textbf{P} & \textbf{R} & \textbf{F1}
& \textbf{P} & \textbf{R} & \textbf{F1}
& \textbf{P} & \textbf{R} & \textbf{F1}
& \textbf{P} & \textbf{R} & \textbf{F1}
& \textbf{P} & \textbf{R} & \textbf{F1}
& \textbf{P} & \textbf{R} & \textbf{F1}
& \textbf{P} & \textbf{R} & \textbf{F1}
& \textbf{P} & \textbf{R} & \textbf{F1} \\
\midrule

Aya-vision-8B & 0.020 & 0.001 & 0.003 & 0.020 & 0.001 & 0.003 & 0.049 & 0.007 & 0.011 & 0.049 & 0.007 & 0.011 & 0.020 & 0.001 & 0.003 & 0.020 & 0.001 & 0.003 & 0.039 & 0.003 & 0.006 & 0.039 & 0.003 & 0.006 \\

Gemma-12B & 0.164 & 0.139 & 0.139 & 0.164 & 0.139 & 0.139 & 0.117 & 0.123 & 0.109 & 0.117 & 0.123 & 0.109 & 0.182 & 0.159 & 0.158 & 0.182 & 0.159 & 0.158 & 0.182 & 0.159 & 0.158 & 0.182 & 0.159 & 0.158 \\

Gemma-4B & 0.130 & 0.123 & 0.114 & 0.130 & 0.123 & 0.114 & 0.120 & 0.118 & 0.107 & 0.120 & 0.118 & 0.107 & 0.096 & 0.088 & 0.082 & 0.096 & 0.088 & 0.082 & 0.096 & 0.088 & 0.082 & 0.096 & 0.088 & 0.082 \\

GPT-4.1 & 0.519 & 0.474 & \textbf{0.475} & 0.519 & 0.474 & \textbf{0.475} &
\textbf{0.333} & \textbf{0.498} & 0.336 & \textbf{0.333} & \textbf{0.498} & 0.336 &
0.316 & \textbf{0.452} & 0.313 & 0.316 & \textbf{0.452} & 0.313 &
0.321 & 0.426 & 0.307 & 0.321 & 0.426 & 0.307 \\

GPT-4.1-mini & \textbf{0.532} & \textbf{0.480} & 0.469 & \textbf{0.532} & \textbf{0.480} & 0.469 &
0.330 & 0.475 & \textbf{0.340} & 0.330 & 0.475 & \textbf{0.340} &
\textbf{0.333} & 0.439 & \textbf{0.322} & \textbf{0.333} & 0.439 & \textbf{0.322} &
\textbf{0.337} & \textbf{0.471} & \textbf{0.340} & \textbf{0.337} & \textbf{0.471} & \textbf{0.340} \\

GPT-4.1-nano & 0.229 & 0.167 & 0.175 & 0.229 & 0.167 & 0.175 &
0.187 & 0.191 & 0.170 & 0.187 & 0.191 & 0.170 &
0.184 & 0.164 & 0.160 & 0.184 & 0.164 & 0.160 &
0.176 & 0.151 & 0.146 & 0.176 & 0.151 & 0.146 \\

GPT-4o & 0.424 & 0.317 & 0.336 & 0.424 & 0.317 & 0.336 &
0.244 & 0.290 & 0.239 & 0.244 & 0.290 & 0.239 &
0.256 & 0.280 & 0.254 & 0.256 & 0.280 & 0.254 &
0.214 & 0.284 & 0.222 & 0.214 & 0.284 & 0.222 \\

GPT-4o-mini & 0.236 & 0.173 & 0.188 & 0.236 & 0.173 & 0.188 &
0.207 & 0.180 & 0.175 & 0.207 & 0.180 & 0.175 &
0.207 & 0.180 & 0.175 & 0.207 & 0.180 & 0.175 &
0.191 & 0.155 & 0.157 & 0.191 & 0.155 & 0.157 \\

LLaMA-3.2-11B & 0.058 & 0.054 & 0.051 & 0.058 & 0.054 & 0.051 &
0.057 & 0.038 & 0.042 & 0.057 & 0.038 & 0.042 &
0.077 & 0.059 & 0.060 & 0.077 & 0.059 & 0.060 &
0.042 & 0.065 & 0.046 & 0.042 & 0.065 & 0.046 \\

Mistral-small-3.1 & 0.424 & 0.412 & 0.395 & 0.424 & 0.412 & 0.395 &
0.261 & 0.421 & 0.299 & 0.261 & 0.421 & 0.299 &
0.262 & 0.395 & 0.295 & 0.262 & 0.395 & 0.295 &
0.233 & 0.361 & 0.267 & 0.233 & 0.361 & 0.267 \\

Pixtral-12B & 0.028 & 0.040 & 0.029 & 0.028 & 0.040 & 0.029 &
0.081 & 0.039 & 0.045 & 0.081 & 0.039 & 0.045 &
0.066 & 0.044 & 0.046 & 0.066 & 0.044 & 0.046 &
0.053 & 0.039 & 0.041 & 0.053 & 0.039 & 0.041 \\

Pixtral-large & 0.402 & 0.285 & 0.277 & 0.402 & 0.285 & 0.277 &
0.173 & 0.232 & 0.151 & 0.173 & 0.232 & 0.151 &
0.245 & 0.137 & 0.132 & 0.245 & 0.137 & 0.132 &
0.237 & 0.217 & 0.148 & 0.237 & 0.217 & 0.148 \\

Qwen-3B & 0.086 & 0.032 & 0.034 & 0.086 & 0.032 & 0.034 &
0.058 & 0.029 & 0.028 & 0.058 & 0.029 & 0.028 &
0.030 & 0.010 & 0.014 & 0.030 & 0.010 & 0.014 &
0.017 & 0.004 & 0.006 & 0.017 & 0.004 & 0.006 \\

Qwen-7B & 0.246 & 0.147 & 0.151 & 0.246 & 0.147 & 0.151 &
0.195 & 0.136 & 0.140 & 0.195 & 0.136 & 0.140 &
0.220 & 0.145 & 0.152 & 0.220 & 0.145 & 0.152 &
0.196 & 0.148 & 0.142 & 0.196 & 0.148 & 0.142 \\

\midrule
\multicolumn{25}{l}{\textbf{Non-VLM Baseline}} \\
BPMN Redrawer & 0.098 & 0.142 & 0.103 & 0.098 & 0.142 & 0.103 & -- & -- & -- & -- & -- & -- & -- & -- & -- & -- & -- & -- & -- & -- & -- & -- & -- & -- \\

Sketch2Process & 0.201 & 0.287 & 0.217 & 0.201 & 0.287 & 0.217 & -- & -- & -- & -- & -- & -- & -- & -- & -- & -- & -- & -- & -- & -- & -- & -- & -- & -- \\

\bottomrule
\end{tabular}
\end{adjustbox}
\label{tab:relations-type-strict-relaxed}
\end{table*}

\begin{table*}[h]
\caption{Strict vs. relaxed evaluation results (types only) with precision, recall, and F1 across VLM-only and OCR-enriched pipelines (best results in bold).}
\centering
\setlength{\tabcolsep}{3pt}
\renewcommand{\arraystretch}{0.9}
\begin{adjustbox}{max width=\textwidth}
\begin{tabular}{lcccccc|cccccc|cccccc|cccccc}
\toprule
\multirow{3}{*}{\textbf{Model}} &
\multicolumn{6}{c|}{\textbf{Only VLM}} &
\multicolumn{6}{c|}{\textbf{VLM + Pix2Struct}} &
\multicolumn{6}{c|}{\textbf{VLM + RapidOCR}} &
\multicolumn{6}{c}{\textbf{VLM + Tesseract}} \\
\cmidrule(lr){2-7} \cmidrule(lr){8-13} \cmidrule(lr){14-19} \cmidrule(lr){20-25}
& \multicolumn{3}{c}{Strict} & \multicolumn{3}{c|}{Relaxed} & \multicolumn{3}{c}{Strict} & \multicolumn{3}{c|}{Relaxed} & \multicolumn{3}{c}{Strict} & \multicolumn{3}{c|}{Relaxed} & \multicolumn{3}{c}{Strict} & \multicolumn{3}{c}{Relaxed} \\
\cmidrule(lr){2-4} \cmidrule(lr){5-7} \cmidrule(lr){8-10} \cmidrule(lr){11-13} \cmidrule(lr){14-16} \cmidrule(lr){17-19} \cmidrule(lr){20-22} \cmidrule(lr){23-25}
& \textbf{P} & \textbf{R} & \textbf{F1} & \textbf{P} & \textbf{R} & \textbf{F1} & \textbf{P} & \textbf{R} & \textbf{F1} & \textbf{P} & \textbf{R} & \textbf{F1} & \textbf{P} & \textbf{R} & \textbf{F1} & \textbf{P} & \textbf{R} & \textbf{F1} & \textbf{P} & \textbf{R} & \textbf{F1} & \textbf{P} & \textbf{R} & \textbf{F1} \\
\midrule
Aya-vision-8B & 0.604 & \textbf{0.929} & 0.715 & 0.704 & \textbf{0.993} & 0.814 & 0.601 & 0.930 & 0.714 & 0.701 & \textbf{1.000} & 0.814 & 0.622 & 0.925 & 0.725 & 0.702 & 0.989 & 0.810 & 0.617 & 0.941 & 0.730 & 0.707 & 0.993 & 0.816 \\
Gemma-12B & 0.683 & 0.585 & 0.617 & 0.707 & 0.675 & 0.683 & 0.617 & 0.643 & 0.614 & 0.686 & 0.693 & 0.680 & 0.719 & 0.801 & 0.743 & 0.790 & 0.846 & 0.808 & 0.719 & 0.801 & 0.743 & 0.790 & 0.846 & 0.808 \\
Gemma-4B & 0.634 & 0.922 & 0.738 & 0.698 & 0.964 & 0.800 & 0.630 & \textbf{0.934} & 0.738 & 0.696 & 0.980 & 0.805 & 0.652 & \textbf{0.957} & 0.761 & 0.710 & \textbf{1.000} & 0.821 & 0.652 & \textbf{0.957} & 0.761 & 0.710 & \textbf{1.000} & 0.821 \\
GPT-4.1 & 0.822 & 0.855 & 0.835 & 0.815 & 0.861 & 0.835 & 0.660 & 0.849 & 0.733 & 0.761 & 0.850 & 0.797 & 0.661 & 0.829 & 0.726 & 0.751 & 0.846 & 0.788 & 0.632 & 0.816 & 0.701 & 0.728 & 0.822 & 0.766 \\
GPT-4.1-mini & 0.839 & 0.878 & \textbf{0.855} & 0.842 & 0.874 & 0.853 & 0.648 & 0.873 & 0.733 & 0.768 & 0.862 & 0.806 & 0.669 & 0.886 & 0.753 & 0.788 & 0.877 & 0.824 & 0.662 & 0.878 & 0.745 & 0.789 & 0.866 & 0.819 \\
GPT-4.1-nano & 0.868 & 0.663 & 0.720 & 0.908 & 0.803 & 0.829 & 0.777 & 0.829 & 0.777 & \textbf{0.880} & 0.937 & 0.895 & 0.765 & 0.813 & 0.770 & \textbf{0.867} & 0.901 & 0.872 & 0.742 & 0.796 & 0.748 & \textbf{0.854} & 0.905 & 0.868 \\
GPT-4o & 0.814 & 0.852 & 0.828 & 0.823 & 0.857 & 0.835 & 0.691 & 0.867 & 0.758 & 0.775 & 0.876 & 0.814 & 0.686 & 0.884 & 0.759 & 0.783 & 0.883 & 0.823 & 0.702 & 0.879 & 0.769 & 0.786 & 0.879 & 0.822 \\
GPT-4o-mini & 0.834 & 0.891 & 0.850 & 0.871 & 0.970 & 0.908 & \textbf{0.785} & 0.928 & \textbf{0.836} & 0.863 & 0.979 & \textbf{0.909} & \textbf{0.785} & 0.928 & \textbf{0.836} & 0.863 & 0.979 & \textbf{0.909} & \textbf{0.773} & 0.927 & \textbf{0.828} & \textbf{0.854} & 0.985 & \textbf{0.907} \\
LLaMA-3.2-11B & 0.580 & 0.912 & 0.685 & 0.663 & 0.958 & 0.766 & 0.602 & 0.915 & 0.700 & 0.660 & 0.951 & 0.761 & 0.597 & 0.913 & 0.697 & 0.675 & 0.965 & 0.781 & 0.573 & 0.901 & 0.674 & 0.652 & 0.951 & 0.755 \\
Mistral-small-3.1 & 0.820 & 0.832 & 0.821 & 0.831 & 0.857 & 0.837 & 0.701 & 0.821 & 0.742 & 0.755 & 0.857 & 0.791 & 0.692 & 0.832 & 0.742 & 0.765 & 0.883 & 0.807 & 0.700 & 0.848 & 0.756 & 0.768 & 0.870 & 0.804 \\
Pixtral-12B & 0.678 & 0.743 & 0.677 & 0.741 & 0.853 & 0.771 & 0.639 & 0.816 & 0.704 & 0.695 & 0.887 & 0.768 & 0.598 & 0.782 & 0.656 & 0.682 & 0.855 & 0.744 & 0.548 & 0.721 & 0.609 & 0.604 & 0.805 & 0.677 \\
Pixtral-large & 0.348 & 0.306 & 0.322 & 0.360 & 0.354 & 0.351 & 0.246 & 0.294 & 0.264 & 0.282 & 0.305 & 0.291 & 0.215 & 0.235 & 0.220 & 0.242 & 0.260 & 0.247 & 0.252 & 0.312 & 0.275 & 0.297 & 0.321 & 0.306 \\
Qwen-3B & 0.447 & 0.730 & 0.544 & 0.533 & 0.765 & 0.619 & 0.437 & 0.692 & 0.526 & 0.525 & 0.745 & 0.608 & 0.242 & 0.380 & 0.290 & 0.286 & 0.412 & 0.334 & 0.288 & 0.447 & 0.343 & 0.337 & 0.471 & 0.388 \\
Qwen-7B & 0.621 & 0.764 & 0.678 & 0.629 & 0.819 & 0.705 & 0.571 & 0.783 & 0.646 & 0.643 & 0.837 & 0.716 & 0.614 & 0.863 & 0.707 & 0.666 & 0.902 & 0.758 & 0.610 & 0.850 & 0.701 & 0.662 & 0.882 & 0.748 \\

\midrule
\multicolumn{25}{l}{\textbf{Non-VLM Baseline}} \\
BPMN Redrawer & 0.931 & 0.806 & \textbf{0.855} & \textbf{0.953} & 0.906 & \textbf{0.921} & -- & -- & -- & -- & -- & -- & -- & -- & -- & -- & -- & -- & -- & -- & -- & -- & -- & -- \\
Sketch2Process & \textbf{0.937} & 0.712 & 0.781 & 0.949 & 0.785 & 0.835 & -- & -- & -- & -- & -- & -- & -- & -- & -- & -- & -- & -- & -- & -- & -- & -- & -- & -- \\

\bottomrule
\end{tabular}
\end{adjustbox}
\label{tab:type-only-strict-relaxed}
\end{table*}

\section{Experiments}
We design our experiments to evaluate how well VLMs can extract structured representations from BPMN diagrams in image form. Models are tested in both vision-only and OCR-enhanced modes, across name, type and relation extraction tasks. 
\subsection{Models Compared}

We benchmark a diverse set of VLMs, including both proprietary APIs and open-source instruction-tuned models hosted on Hugging Face\footnote{\url{https://huggingface.co/}} \cite{wolf2020transformers}. Our selection spans a wide range of architectures and capabilities: 
\textbf{GPT-4.1 (standard, mini, nano)}\footnote{\url{https://openai.com/index/gpt-4-1/}} and \textbf{GPT-4o (standard, mini)} \cite{hurst2024gpt} from OpenAI; 
\textbf{Qwen2.5-VL models (7B and 3B)} \cite{qwen2.5-VL,Qwen-VL} from Alibaba group; 
\textbf{Pixtral-12B} and \textbf{Pixtral-Large} \cite{agrawal2024pixtral} and 
\textbf{Mistral-Small-3.1}\footnote{\url{https://mistral.ai/news/mistral-small-3-1}} from MistralAI team; 
\textbf{Aya-Vision-8B}\footnote{\url{https://cohere.com/blog/aya-vision}} from CohereForAI; 
\textbf{Gemma3-4B and 12B} \cite{team2025gemma} from Google; and 
\textbf{LLaMA-3.2-11B-Vision-Instruct} \cite{meta2024llama3.2-vision} from Meta.
 This range allows a comprehensive analysis of multimodal diagram understanding across model scales and training paradigms. 

Each model is evaluated in two configurations:
\begin{itemize}
  \item \textbf{Vision-only}: The model predicts component names and types directly from the BPMN image.
  \item \textbf{OCR-enhanced}: OCR text is appended only when the model fails to detect key components such as events, gateways or sequence flows. 
\end{itemize}

This allows us to assess both the raw visual understanding of the model and its ability to incorporate OCR-enriched information for better name and type extraction.

In addition, we include \textbf{BPMN Redrawer} \cite{antinori2022bpmn} and \textbf{Sketch2\-Process} \cite{schafer2022sketch2process} as non-VLM baselines. \textbf{BPMN Redrawer}\footnote{\url{https://github.com/PROSLab/BPMN-redrawer}}  converts BPMN images into executable diagrams by combining convolutional neural networks with OCR-based text recognition, while \textbf{Sketch2Process}\footnote{\url{https://github.com/dwslab/hdBPMN}} translates hand-drawn sketches into BPMN models using deep learning–based symbol recognition.

We evaluate performance using Precision, Recall, and F1 Score, applying multiple levels of matching criteria to capture both exactness and tolerance to partial or noisy predictions. Gold-standard CSV files are constructed by parsing the corresponding \texttt{.bpmn} XML files for each diagram. These files provide the name, type, and relations of each BPMN element and serve as the reference for evaluation. In the name-only setting, predicted element names are compared with gold-standard names regardless of type. The type-only setting focuses exclusively on element categories, abstracting away from labels. The name+type setting requires both fields to be present and identical, with incomplete entries excluded to ensure fairness. Finally, the relations setting evaluates whether both the type and connectivity of elements align with the gold standard.  

\section{Results and Analysis}
We evaluate model performance under four evaluation regimes: \emph{name-only}, \emph{name+type}, \emph{relations+type}, and \emph{type-only}. Each regime is measured in both strict and relaxed variants where applicable, comparing VLM-only inference with OCR-enriched pipelines. The non-VLM baseline is only reported without enrichment. In the relaxed settings, type labels may partially match (e.g., \texttt{exclusivegateway} $\approx$ \texttt{gateway}). Model outputs are recorded as either \texttt{.json} (valid structured output) or \texttt{.txt} (unstructured output requiring post-processing). In particular, LLaMA-3.2-11B often produces \texttt{.txt} responses, requiring additional parsing before evaluation.


Formally, let $G$ be the gold-standard set of elements and $M$ the model predictions.
Each element has a name $n$, a type $t$, or a relation $r=(s,d,t)$.
Matching is defined as follows:
\begin{equation}
M_{\text{name}}(n_g,n_m) =
\begin{cases}
1 & \text{if } n_g = n_m, \\
0 & \text{otherwise}.
\end{cases}
\end{equation}

\begin{equation}
M_{\text{type}}(t_g,t_m) =
\begin{cases}
1 & \text{if } t_g = t_m \quad \text{(strict)}, \\
1 & \text{if } \mathrm{norm}(t_g)=\mathrm{norm}(t_m) \quad \text{(relaxed)}, \\
0 & \text{otherwise}.
\end{cases}
\end{equation}

\begin{equation}
M_{\text{name-type}}((n_g,t_g),(n_m,t_m)) =
\begin{cases}
1 & \text{if } n_g=n_m \wedge M_{\text{type}}^{\ast}(t_g,t_m)=1, \\
0 & \text{otherwise},
\end{cases}
\end{equation}
where $M_{\text{type}}^{\ast}$ denotes either strict or relaxed type matching.

\begin{equation}
\begin{aligned}
M_{\text{rel-type}}((s_g,d_g,t_g),(s_m,d_m,t_m)) =
\begin{cases}
1 & \text{if } (s_g,d_g)=(s_m,d_m) \\
  & \;\;\;\;\;\wedge\; M_{\text{type}}^{\ast}(t_g,t_m)=1, \\
0 & \text{otherwise},
\end{cases}
\end{aligned}
\end{equation}
with $M_{\text{type}}^{\ast}$ again being strict or relaxed.

True/false sets under regime $X$ are:
\begin{equation}
\begin{aligned}
TP_X &= \{ g \in G \mid \exists m \in M: M_X(g,m)=1 \}, \\
FP_X &= \{ m \in M \mid \nexists g \in G: M_X(g,m)=1 \}, \\
FN_X &= \{ g \in G \mid \nexists m \in M: M_X(g,m)=1 \}.
\end{aligned}
\end{equation}

These definitions provide the basis for computing Precision, Recall, and F1 scores under each evaluation regime.

\subsection{Name-Only Evaluation (Table~\ref{tab:names-only-strict})}
Focusing solely on element names, this evaluation highlights text grounding capabilities.

\textbf{Top-tier models} such as GPT-4.1, GPT-4o, and Mistral-Small-3.1 achieve the highest F1 scores ($>$0.70). They perform robustly without OCR, with enrichment providing minimal or even negative gains due to noise. \textbf{Mid-tier models}, including Qwen2.5VL-7B, Gemma-12B, and Pixtral-Large, benefit more from OCR. Tesseract and Pix2Struct improve recall, helping these models recover missing labels at a modest precision cost. \textbf{Low-tier models}, such as Qwen2.5VL-3B, Aya-Vision-8B, and Pixtral-12B, show little improvement or degrade with OCR. Their limited multimodal integration leaves them vulnerable to spurious OCR tokens. Finally, the \textbf{non-VLM baselines} perform competitively in name-only evaluation, underscoring the continuing value of traditional vision+OCR pipelines alongside VLMs.

\subsection{Name+Type Evaluation (Table~\ref{tab:strict-relaxed-eval})}
This evaluation demands correct extraction of both names and types, reported under strict and relaxed variants.

\textbf{Strict setting:}  
\textbf{Top-tier models} 
achieve the best results, though their F1 drops relative to name-only evaluation. GPT-4.1-mini slightly outperforms GPT-4.1, suggesting stronger calibration on type distinctions. OCR brings little benefit at this tier.  
\textbf{Mid-tier models} 
benefit substantially from OCR, particularly Pix2Struct and Tesseract, which help recover fine-grained type labels missed in vision-only inference.  
\textbf{Low-tier models} 
remain weak; noisy OCR often harms their performance further. The \textbf{non-VLM baselines} also perform competitively here, with \textbf{Sketch2Process} in particular rivaling mid-tier VLMs; their advantage comes from specialized CNN+OCR pipelines tuned for BPMN structure, which compensate for the lack of multimodal reasoning.

\textbf{Relaxed setting:}  
Allowing partial type matches improves recall across all models. \textbf{Top-tier models} remain stable, showing minimal OCR gains, consistent with strong inherent multimodal grounding. \textbf{Mid-tier models} gain most (+3–5 F1), especially Pixtral-Large and Gemma-12B with Pix2Struct and Tesseract. \textbf{Low-tier models} continue to underperform, with OCR providing little or no benefit.

\subsection{Relations+Type Evaluation (Table~\ref{tab:relations-type-strict-relaxed})}
This more demanding evaluation requires both correct relation extraction and accurate typing.

\textbf{Strict setting:}  
\textbf{Top-tier models} 
maintain the highest scores, although relations prove harder to capture than names and types alone. OCR adds limited value at this tier.  
\textbf{Mid-tier models} clearly benefit from OCR, with Tesseract and Pix2Struct aiding recovery of relation labels and directionality.  
\textbf{Low-tier models} struggle significantly, with OCR often worsening results by injecting spurious connections. The \textbf{non-VLM baselines} lag behind VLMs, as their handcrafted CNN+OCR pipelines are tuned for symbol and text recognition but lack semantic grounding, making relation directionality especially error-prone.

\textbf{Relaxed setting:}  
Relaxed matching improves recall noticeably, especially for mid-tier models. Pixtral-Large and Gemma-12B show gains of several F1 points under OCR. High-tier models remain strong and OCR-agnostic, while low-tier models remain unreliable, with noisy enrichment outweighing any benefit.

\subsection{Type-Only Evaluation (Table~\ref{tab:type-only-strict-relaxed})}
Finally, we evaluate models on type classification alone.

\textbf{Strict setting:}  
\textbf{Top-tier models} 
maintain strong type discrimination, relying primarily on visual shape recognition. \textbf{Mid-tier models} 
benefit from OCR, particularly Pix2Struct, which helps with ambiguous event and gateway types. \textbf{Low-tier models} continue to underperform, with OCR introducing additional errors. The \textbf{non-VLM baselines} achieve competitive results here because type recognition is largely a visual classification task; their CNN+OCR pipelines are highly optimized for symbol shapes, giving them an advantage over weaker VLMs.

\textbf{Relaxed setting:}  
Relaxed scoring boosts recall across models. \textbf{Top-tier systems} remain relatively unaffected by OCR. \textbf{Mid-tier models} show the greatest improvements (up to +4–5 F1), especially when enriched with Pix2Struct and Tesseract. \textbf{Low-tier models} remain weak, with OCR contributing little value.

\subsection{Error Distribution Across Element Types} \label{sec:error_dist}
To complement the aggregate metrics, we analyzed which BPMN element types were more prone to errors across models. Table~\ref{tab:error_rates_overall} summarizes the overall error rates normalized by the frequency of each element type in the gold-standard annotations. 

\begin{table}[h]
\centering
\small
\renewcommand{\arraystretch}{0.9}
\caption{Top 10 BPMN element types with highest error rates across models.}
\label{tab:error_rates_overall}
\begin{tabular}{lccc}
\toprule
          Element Type &  Gold Count &  Error Count &  Error Rate \\
\midrule
             datastore &           1 &          202 &     202.000 \\
          sequenceflow &         482 &        66958 &     138.917 \\
           messageflow &         136 &        15358 &     112.926 \\
intermediatecatchevent &           1 &          106 &     106.000 \\
            startevent &          97 &        10188 &     105.031 \\
     intermediateevent &         125 &        13020 &     104.160 \\
                  pool &          98 &         9376 &      95.673 \\
                 event &           1 &           85 &      85.000 \\
      exclusivegateway &         106 &         8616 &      81.283 \\
              endevent &          99 &         7605 &      76.818 \\
\bottomrule
\end{tabular}
\end{table}

The results indicate that certain control-flow constructs, such as \emph{sequence flows}, \emph{message flows}, and \emph{gateways}, are consistently more challenging for models, yielding disproportionately high error rates compared to their gold frequency. 
This arises because these elements depend on correctly modeling relational structure and directional semantics, which VLMs struggle to infer from visual cues alone. Conversely, simpler constructs like \emph{tasks} and \emph{events} also show substantial error counts due to their prevalence, though their normalized error rates are somewhat lower. Their errors typically stem from noisy OCR labels and frequent visual overlaps, rather than structural ambiguity.

\section{Statistical Analysis}

To complement the raw evaluation scores, we conducted statistical significance testing to assess whether differences across OCR settings and models were robust. We combined non-parametric tests with effect size reporting to provide both inferential and practical insights. 
Specifically, this section addresses the following research questions:

\begin{itemize}
    \item \textbf{RQ1:} Are the observed improvements from OCR statistically significant across models, and do certain models benefit more than others?
    \item \textbf{RQ2:} How consistent are model performances across OCR settings, and what are the relative rankings of models when evaluated jointly across all conditions?
    \item \textbf{RQ3:} What is the practical magnitude of the impact of OCR on the performance of the model? 
\end{itemize}

\subsection{Pairwise OCR vs. VLM-only Comparisons}
We first compared each OCR-enriched pipeline (Pix2Struct, RapidOCR, and Tesseract) against the VLM-only baseline for each model using the Wilcoxon signed-rank \cite{wilcoxon1945individual} test. Results in Table~\ref{tab:pairwise-full} show that top-performing models (e.g., GPT-4.1, GPT-4o, GPT-4.1-mini, and Mistral-Small-3.1) exhibit highly significant differences ($p < 0.01$) across all OCR methods. Mid-range models such as Qwen2.5-7B show selective gains, while weaker models (e.g., Aya-Vision, Pixtral-12B) show no significant improvements. This directly answers \textbf{RQ1}, showing that OCR benefits are model-dependent.

\begin{table}[h]
\centering
\caption{Pairwise Wilcoxon test results (OCR vs VLM-only, F1). 
Lower $p$-values indicate stronger evidence of significant differences.}
\label{tab:pairwise-full}
\begin{adjustbox}{max width=\columnwidth}
\begin{tabular}{lccc}
\toprule
\textbf{Model} & \textbf{VLM vs Pix2Struct} & 
                 \textbf{VLM vs RapidOCR} & 
                 \textbf{VLM vs Tesseract} \\
\midrule
Aya-Vision-8B & 0.945 & 0.383 & 0.203 \\
GPT-4.1 & 0.008 & 0.008 & 0.008 \\
GPT-4.1-mini & 0.008 & 0.008 & 0.008 \\
GPT-4.1-nano & 0.078 & 0.078 & 0.078 \\
GPT-4o & 0.008 & 0.008 & 0.008 \\
GPT-4o-mini & 0.008 & 0.008 & 0.008 \\
Gemma-4B & 0.883 & 0.883 & 0.883 \\
Gemma-12B & 0.008 & 0.008 & 0.008 \\
LLaMA-3.2-11B & 0.453 & 0.453 & 0.453 \\
Mistral-Small-3.1 & 0.008 & 0.008 & 0.008 \\
Pixtral-12B & 0.195 & 0.195 & 0.195 \\
Pixtral-Large & 0.008 & 0.008 & 0.008 \\
Qwen2.5-3B & 0.305 & 0.305 & 0.305 \\
Qwen2.5-7B & 0.016 & 0.016 & 0.016 \\
\bottomrule
\end{tabular}
\end{adjustbox}
\end{table}

\subsection{Model Consistency Across OCR Settings}
To evaluate consistency across OCR conditions within each model, we applied the Friedman test \cite{friedman1937use} with Kendall’s \cite{kendall1939problem} $W$ as an effect size as shown in Table~\ref{tab:friedman-full}. High $W$ values (e.g., GPT-4.1: $W=1.0$; Pixtral-Large: $W=0.889$; Gemma-12B: $W=0.854$) indicate strong agreement that OCR systematically shifts performance. In contrast, models such as Aya-Vision and LLaMA-3.2-11B yield low $W$, suggesting random variability rather than systematic improvements. These findings address \textbf{RQ2}, highlighting which models are robust across OCR settings.

\begin{table}[h]
\small
\centering
\renewcommand{\arraystretch}{0.9}
\caption{Friedman test results across OCR settings (F1). 
Lower $p$-values ($\downarrow$) indicate significance; higher Kendall’s $W$ ($\uparrow$) indicates stronger agreement ($W \in [0,1]$).}
\label{tab:friedman-full}
\begin{tabular}{lcc}
\toprule
\textbf{Model} & \textbf{$p$-value ($\downarrow$)} & \textbf{Kendall’s $W$ ($\uparrow$)} \\
\midrule
Aya-Vision-8B & 0.648 & 0.069 \\
Gemma-4B & 0.881 & 0.028 \\
Gemma-12B & 0.000 & 0.854 \\
GPT-4.1 & 0.000 & 1.000 \\
GPT-4.1-mini & 0.000 & 0.775 \\
GPT-4.1-nano & 0.007 & 0.506 \\
GPT-4o & 0.002 & 0.625 \\
GPT-4o-mini & 0.007 & 0.471 \\
LLaMA-3.2-11B & 0.444 & 0.083 \\
Mistral-Small-3.1 & 0.000 & 0.791 \\
Pixtral-12B & 0.444 & 0.083 \\
Pixtral-Large & 0.000 & 0.889 \\
Qwen2.5-3B & 0.305 & 0.139 \\
Qwen2.5-7B & 0.037 & 0.292 \\
\bottomrule
\end{tabular}
\end{table}

\subsection{Model Ranking Across Settings}
\begin{figure}[h]
    \centering
    \includegraphics[width=\linewidth]{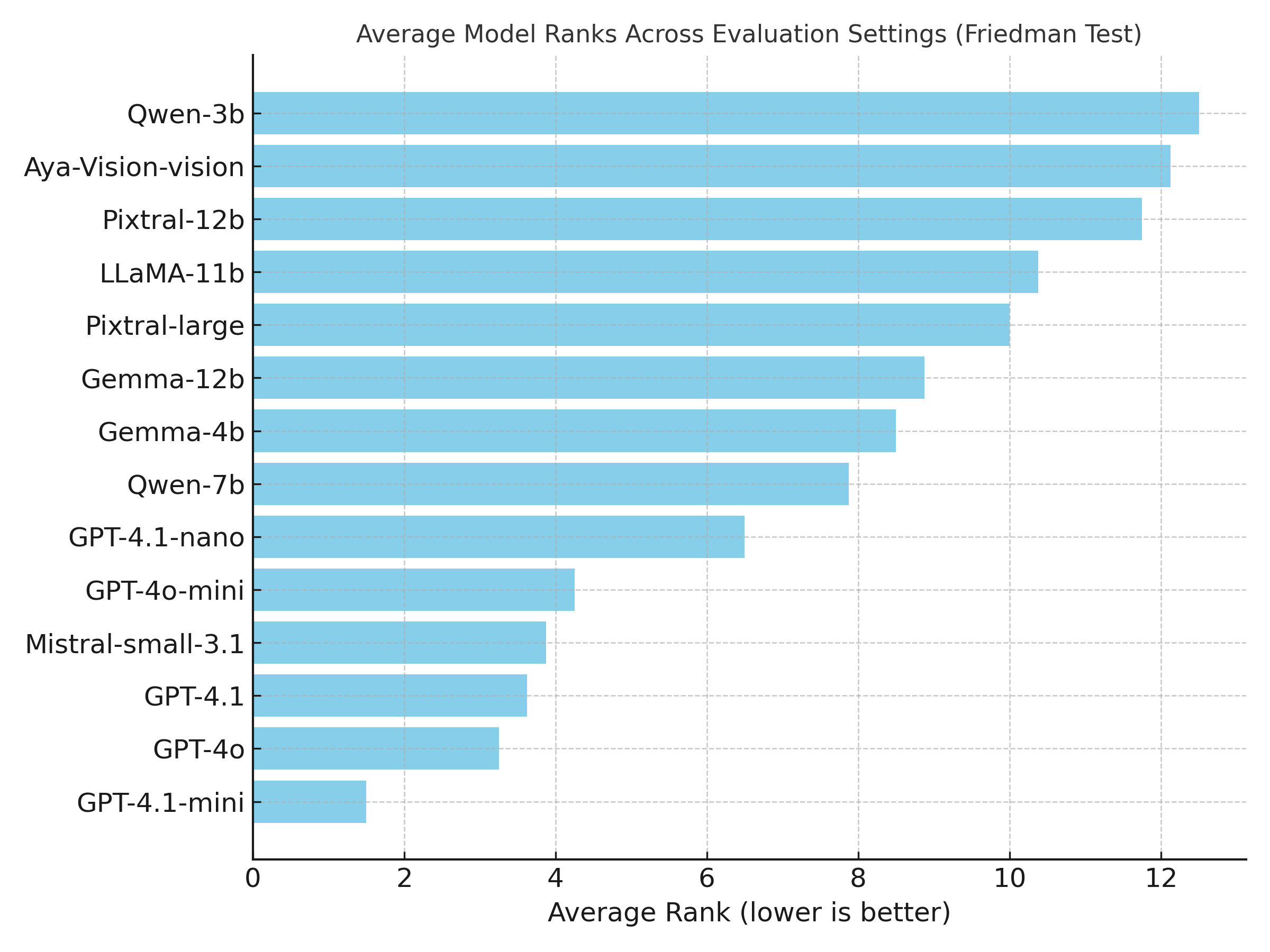}
    \caption{Average model ranks across evaluation settings based on the Friedman test. Lower ranks indicate better performance.}
    \label{fig:avg-ranks}
\end{figure}

The Friedman framework also allows us to rank models by average performance across all evaluation settings. Figure~\ref{fig:avg-ranks} illustrates that GPT-4.1, GPT-4o, GPT-4.1-mini, and Mistral-Small-3.1 consistently occupy the top ranks, while Pixtral-12B and Aya-Vision remain at the bottom. This ranking provides a global perspective on robustness independent of individual evaluation metrics and further answers \textbf{RQ2}.

\subsection{Effect Sizes of OCR Impact}
\begin{figure}[h]
    \centering
    \includegraphics[width=\linewidth]{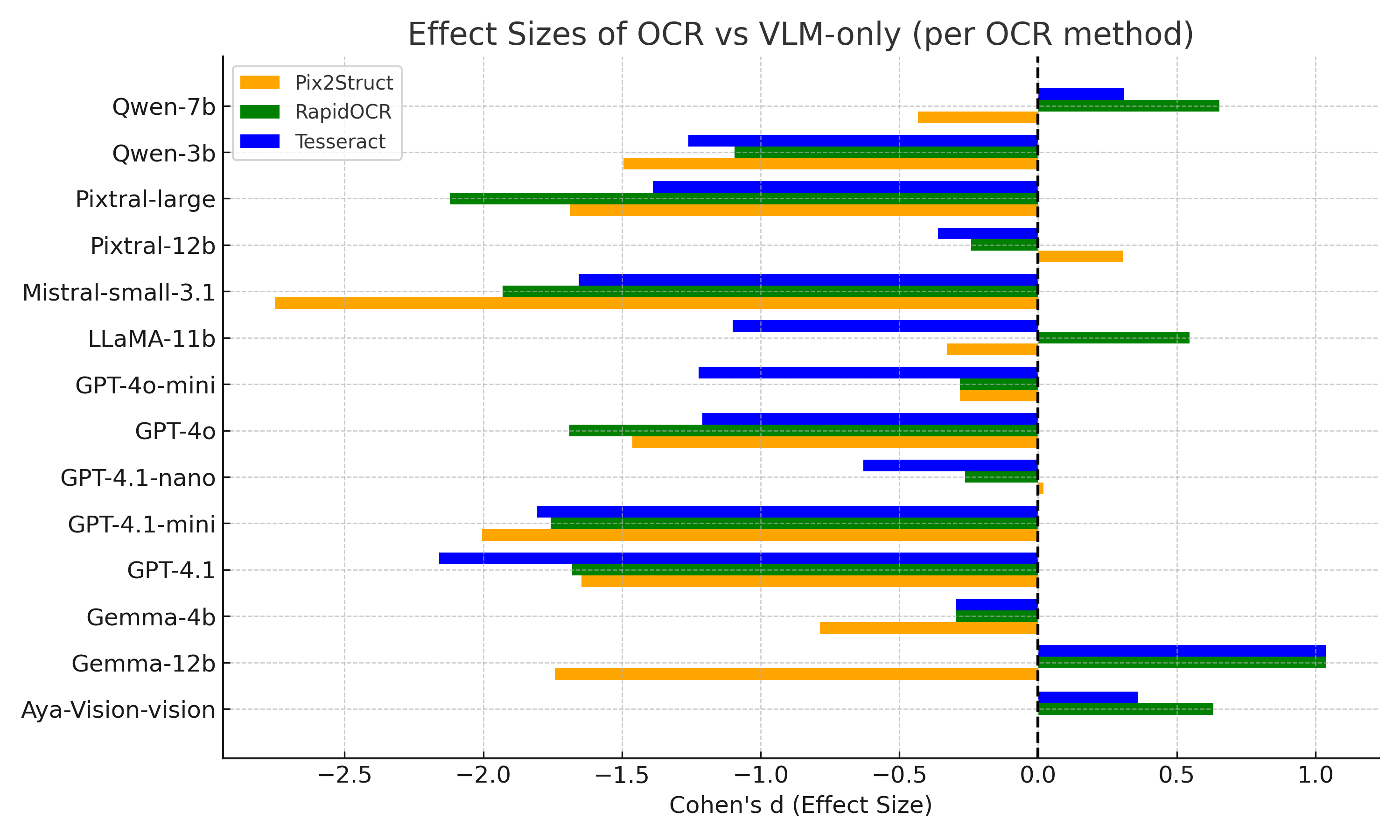}
    \caption{Effect sizes (Cohen's $d$) for OCR vs. VLM-only across models and OCR methods. Positive values indicate performance gains from OCR.}
    \label{fig:effect-sizes}
\end{figure}

While significance tests establish whether OCR helps, effect sizes quantify ``how much'' it helps. Figure~\ref{fig:effect-sizes} 
present Cohen’s $d$ \cite{cohen2013statistical} for each model and OCR method. Negative values indicate degradation with OCR, while positive values indicate improvements. Top-tier models (e.g., GPT-4.1, GPT-4o, Mistral-Small-3.1) show consistently large negative $d$ values, reflecting that OCR may inject noise into otherwise strong visual extraction. By contrast, mid-tier models (e.g., Qwen2.5-7B, Gemma-12B) yield positive $d$, confirming that they benefit most from additional textual cues. This analysis directly addresses \textbf{RQ3} by moving beyond statistical significance to practical significance.

\section{Ablation Study on Prompt Variants}

To assess the impact of prompt design on BPMN extraction, we evaluated four variants\footnote{The prompt variations can be accessed via \url{https://github.com/pritamdeka/BPMN-VLM/tree/main/prompts/prompt_ablations}} of the baseline schema, ranging from minimal instructions to reasoning-augmented and traversal-based strategies as described in Table~\ref{tab:prompt-ablation}. Experiments were conducted across four evaluation settings: \textbf{Name-only}, \textbf{Type-only}, \textbf{Name+Type (Entities)}, and \textbf{Relation+Type (Relations)}. Table~\ref{tab:full-strict-all} reports strict F1 scores for each model and strategy, while Table~\ref{sst-prompt-types} presents Wilcoxon significance tests, highlighting which prompt types yield consistent improvements.

\begin{table}[h]
\centering
\small
\caption{Prompting variants explored in the ablation study for BPMN diagram extraction.}
\begin{adjustbox}{max width=\columnwidth}
\renewcommand{\arraystretch}{0.95}
\begin{tabular}{|>{\columncolor{gray!10}}p{0.20\linewidth}|p{0.73\linewidth}|}
\hline
\textbf{Variant} & \textbf{Description} \\
\hline
\textbf{Baseline} &
Original prompt with explicit instructions for extracting BPMN elements, bounding boxes, labels, and flows, along with an example JSON schema. \\
\hline
\textbf{Only-Example} &
Removes explicit instructions and supplies only the example JSON schema. This condition examines if the model can deduce the extraction strategy implicitly from the example. \\
\hline
\textbf{Chain-of-Thought (CoT)} &
Adds hidden reasoning instructions: the model is asked to think step-by-step silently (scratchpad) but only emit the final JSON output. Ensures more systematic parsing while preventing verbose outputs. \\
\hline
\textbf{Self-Consistency} &
Extends CoT by requiring the model to internally generate multiple candidate outputs and select the most consistent one. Implemented as best-of-$3$ decoding to reduce random errors and inconsistencies. \\
\hline
\textbf{DFS+BFS Hybrid} &
Augments the baseline with traversal strategy: first apply Breadth-First Search (BFS) to capture overall structure (pools, lanes, tasks, events), then Depth-First Search (DFS) to follow each flow from start to end. Balances structural coverage with flow completeness. \\
\hline
\end{tabular}
\end{adjustbox}
\label{tab:prompt-ablation}
\end{table}

\paragraph{Analysis of Prompt Strategies.} 
Table~\ref{tab:full-strict-all} shows that performance varies considerably by prompt type and model. The baseline prompt provides stable but modest scores across all evaluation categories. The Only-Example prompt consistently underperforms, confirming that example schemas alone are insufficient for robust diagram parsing. Chain-of-Thought and Self-Consistency generally yield incremental gains for models such as GPT-4.1, GPT-4o, Gemma-12B, suggesting that reasoning and ensemble-style decoding help capture complex control-flow constructs. The DFS+BFS hybrid prompt stands out as the most effective in many cases, especially for entity and relation extraction, as it enforces structural traversal of diagrams and reduces missed flows. Differences across models also highlight capacity effects: larger models exploit advanced prompting more effectively, while smaller models show only modest gains, often limited by representational power rather than prompting strategy.

\begin{table*}[h]
\centering
\small
\setlength{\tabcolsep}{3pt}
\caption{Strict evaluation results (F1 scores) across models and prompt strategies \textbf{(best results in bold)}.}
\label{tab:full-strict-all}
\begin{adjustbox}{max width=\textwidth}
\begin{tabular}{|l|cccc|cccc|cccc|cccc|cccc|}
\hline
\multirow{2}{*}{\textbf{Model}}
& \multicolumn{4}{c|}{\textbf{Baseline}}
& \multicolumn{4}{c|}{\textbf{Only-Example}}
& \multicolumn{4}{c|}{\textbf{Chain-of-Thought}}
& \multicolumn{4}{c|}{\textbf{Self-Consistency}}
& \multicolumn{4}{c|}{\textbf{DFS+BFS}} \\
\cline{2-21}
& \textbf{Ent.} & \textbf{Rel.} & \textbf{Name} & \textbf{Type}
& \textbf{Ent.} & \textbf{Rel.} & \textbf{Name} & \textbf{Type}
& \textbf{Ent.} & \textbf{Rel.} & \textbf{Name} & \textbf{Type}
& \textbf{Ent.} & \textbf{Rel.} & \textbf{Name} & \textbf{Type}
& \textbf{Ent.} & \textbf{Rel.} & \textbf{Name} & \textbf{Type} \\
\hline
Aya-Vision & 0.11 & 0.00 & 0.05 & 0.04 & 0.10 & 0.00 & 0.05 & 0.04 & 0.10 & \textbf{0.01} & 0.05 & 0.04 & 0.10 & 0.00 & 0.05 & 0.04 & \textbf{0.12} & 0.00 & \textbf{0.06} & \textbf{0.05} \\
\hline
Gemma-12B & 0.42 & 0.14 & 0.21 & 0.22 & \textbf{0.58} & 0.19 & 0.27 & 0.29 & 0.54 & 0.11 & 0.26 & 0.28 & 0.56 & 0.13 & 0.27 & 0.29 & 0.57 & \textbf{0.21} & \textbf{0.28} & \textbf{0.30} \\
\hline
Gemma-4B & 0.43 & \textbf{0.11} & 0.18 & \textbf{0.19} & 0.41 & 0.09 & 0.18 & 0.18 & \textbf{0.46} & 0.09 & \textbf{0.19} & \textbf{0.19} & 0.45 & 0.08 & 0.18 & \textbf{0.19} & 0.44 & 0.10 & \textbf{0.19} & \textbf{0.19} \\
\hline
GPT-4.1 & 0.56 & 0.22 & 0.30 & 0.31 & 0.71 & 0.26 & 0.36 & \textbf{0.38} & 0.70 & 0.27 & 0.36 & 0.37 & 0.70 & 0.24 & 0.35 & 0.37 & \textbf{0.72} & \textbf{0.29} & \textbf{0.37} & \textbf{0.38} \\
\hline
GPT-4.1-mini & 0.54 & 0.21 & 0.29 & 0.30 & 0.66 & 0.24 & 0.34 & 0.35 & 0.67 & 0.23 & 0.34 & 0.35 & 0.67 & 0.22 & 0.34 & 0.35 & \textbf{0.68} & \textbf{0.26} & \textbf{0.35} & \textbf{0.36} \\
\hline
GPT-4.1-nano & 0.33 & 0.09 & 0.17 & 0.18 & 0.42 & 0.11 & 0.20 & 0.21 & 0.42 & 0.10 & 0.20 & 0.21 & 0.41 & 0.10 & 0.19 & 0.20 & \textbf{0.43} & \textbf{0.13} & \textbf{0.21} & \textbf{0.22} \\
\hline
GPT-4o & 0.57 & 0.24 & 0.31 & 0.32 & 0.71 & 0.28 & 0.37 & 0.38 & 0.71 & 0.29 & 0.37 & 0.38 & 0.72 & 0.26 & 0.36 & 0.38 & \textbf{0.73} & \textbf{0.30} & \textbf{0.38} & \textbf{0.39} \\
\hline
GPT-4o-mini & 0.55 & 0.22 & 0.30 & 0.31 & 0.67 & 0.25 & 0.35 & 0.36 & 0.68 & 0.25 & 0.35 & 0.36 & 0.68 & 0.23 & 0.35 & 0.36 & \textbf{0.69} & \textbf{0.27} & \textbf{0.36} & \textbf{0.37} \\
\hline
LLaMA-11B & 0.38 & 0.11 & 0.20 & 0.21 & 0.48 & 0.13 & 0.23 & 0.24 & 0.48 & 0.12 & 0.23 & 0.24 & 0.47 & 0.11 & 0.23 & 0.24 & \textbf{0.49} & \textbf{0.14} & \textbf{0.24} & \textbf{0.25} \\
\hline
Mistral-Small-3.1 & 0.35 & 0.10 & 0.18 & 0.19 & 0.43 & 0.12 & 0.20 & 0.21 & 0.44 & 0.11 & \textbf{0.21} & 0.21 & 0.44 & 0.11 & 0.20 & 0.21 & \textbf{0.45} & \textbf{0.13} & \textbf{0.21} & \textbf{0.22} \\
\hline
Pixtral-12B & 0.42 & 0.16 & 0.22 & 0.23 & 0.53 & 0.18 & 0.26 & 0.27 & 0.53 & 0.18 & 0.26 & 0.27 & 0.53 & 0.17 & 0.26 & 0.27 & \textbf{0.54} & \textbf{0.20} & \textbf{0.27} & \textbf{0.28} \\
\hline
Pixtral-Large & 0.44 & 0.18 & 0.23 & 0.24 & 0.56 & 0.20 & 0.28 & 0.29 & 0.56 & 0.20 & 0.28 & 0.29 & 0.56 & 0.19 & 0.27 & 0.29 & \textbf{0.57} & \textbf{0.22} & \textbf{0.29} & \textbf{0.30} \\
\hline
Qwen-3B & 0.31 & 0.08 & 0.15 & 0.16 & 0.39 & 0.09 & 0.18 & 0.19 & 0.39 & 0.09 & 0.18 & 0.19 & 0.39 & 0.09 & 0.18 & 0.19 & \textbf{0.40} & \textbf{0.10} & \textbf{0.19} & \textbf{0.20} \\
\hline
Qwen-7B & 0.37 & 0.11 & 0.19 & 0.20 & 0.47 & 0.13 & 0.22 & 0.23 & 0.47 & 0.12 & 0.22 & 0.23 & 0.47 & 0.12 & 0.22 & 0.23 & \textbf{0.48} & \textbf{0.14} & \textbf{0.23} & \textbf{0.24} \\
\hline
\end{tabular}
\end{adjustbox}
\end{table*}

\paragraph{Statistical Significance Test}

The significance test results from Table~\ref{sst-prompt-types} show that prompt strategies which reduce ambiguity, such as the \textbf{Baseline} and especially the \textbf{DFS+BFS hybrid}, consistently outperform alternatives. Baseline works reliably by pairing explicit instructions with an example schema, while DFS+BFS is most effective because its traversal policy mirrors the logical structure of BPMN diagrams, reducing missed dependencies and improving relational accuracy. In contrast, \textbf{Only-Example} performs weakest, as models fail to infer extraction rules from a schema alone. Reasoning-heavy strategies yield mixed outcomes: \textbf{Chain-of-Thought} adds limited benefit since BPMN extraction depends more on explicit mapping than abstract reasoning, and \textbf{Self-Consistency} mainly helps mid-sized models by reducing stochastic errors. Model-wise, \textbf{large models} such as GPT-4.1, GPT-4o benefit most from structured traversal, \textbf{mid-sized models} (Gemma-12B, Qwen-7B, Pixtral-12B) gain moderately from both DFS+BFS and Self-Consistency, while \textbf{smaller models} (GPT-4.1-mini/nano, Gemma-4B, Qwen-3B) are most sensitive, failing under Only-Example and CoT but performing well for DFS+BFS.

\begin{table}[h]
\centering
\setlength{\tabcolsep}{4pt}
\caption{Wilcoxon significance test results (p-values) for Name+Type and Relations+Type across prompt types \textbf{(lowest p-values in bold)}.}
\begin{adjustbox}{max width=\columnwidth}
\begin{tabular}{l|cc|cc|cc|cc|cc}
\toprule
\multirow{2}{*}{\textbf{Model}} 
  & \multicolumn{2}{c|}{\textbf{Baseline}} 
  & \multicolumn{2}{c|}{\textbf{COT}} 
  & \multicolumn{2}{c|}{\textbf{DFS-BFS}} 
  & \multicolumn{2}{c|}{\textbf{Only\_example}} 
  & \multicolumn{2}{c}{\textbf{Self\_consistency}} \\
\cmidrule(lr){2-3} \cmidrule(lr){4-5} \cmidrule(lr){6-7} \cmidrule(lr){8-9} \cmidrule(lr){10-11}
 & \textbf{Ent} & \textbf{Rel} 
 & \textbf{Ent} & \textbf{Rel} 
 & \textbf{Ent} & \textbf{Rel} 
 & \textbf{Ent} & \textbf{Rel} 
 & \textbf{Ent} & \textbf{Rel} \\
\midrule
Aya-Vision & 0.74 & 0.29 & 0.76 & 0.21 & \textbf{0.44} & \textbf{0.09} & 0.77 & \textbf{0.09} & 0.75 & 0.32 \\
Gemma-12B & \textbf{0.01} & 0.70 & 0.36 & 0.30 & 0.43 & \textbf{0.29} & 0.41 & 0.31 & 0.48 & 0.42 \\
Gemma-4B  & 0.39 & 0.28 & 0.54 & 0.49 & 0.55 & 0.41 & \textbf{0.26} & \textbf{0.14} & 0.56 & 0.54 \\
GPT-4.1   & 0.48 & 0.70 & \textbf{0.44} & 0.68 & 0.59 & 0.78 & 0.63 & 0.67 & 0.59 & \textbf{0.59} \\
GPT-4.1-mini & 0.69 & \textbf{0.83} & 0.72 & 0.85 & 0.73 & 0.91 & 0.70 & 0.85 & \textbf{0.68} & 0.84 \\
GPT-4.1-nano & 0.71 & \textbf{0.79} & 0.67 & 0.82 & \textbf{0.64} & 0.91 & 0.65 & 0.80 & 0.66 & 0.81 \\
GPT-4o    & \textbf{0.47} & 0.66 & 0.48 & \textbf{0.65} & 0.52 & 0.72 & 0.51 & 0.69 & 0.53 & 0.70 \\
GPT-4o-mini & 0.62 & 0.76 & 0.61 & 0.77 & \textbf{0.58} & \textbf{0.72} & 0.60 & 0.73 & 0.59 & 0.74 \\
LLaMA-11B & 0.44 & 0.60 & \textbf{0.42} & \textbf{0.58} & 0.45 & 0.61 & 0.43 & 0.59 & 0.46 & 0.62 \\
Mistral-Small-3.1 & 0.54 & 0.72 & \textbf{0.53} & \textbf{0.71} & 0.55 & 0.73 & 0.56 & 0.74 & 0.55 & 0.73 \\
Pixtral-12B & 0.40 & 0.63 & \textbf{0.39} & \textbf{0.62} & 0.41 & 0.64 & 0.42 & 0.65 & 0.41 & 0.64 \\
Pixtral-Large & \textbf{0.46} & \textbf{0.68} & 0.47 & 0.69 & 0.48 & 0.70 & 0.49 & 0.71 & 0.47 & 0.69 \\
Qwen-3B   & \textbf{0.50} & \textbf{0.72} & 0.51 & 0.73 & 0.52 & 0.74 & 0.53 & 0.75 & 0.51 & 0.73 \\
Qwen-7B   & \textbf{0.44} & \textbf{0.65} & 0.45 & 0.66 & 0.47 & 0.68 & 0.46 & 0.67 & 0.45 & 0.66 \\
\bottomrule
\end{tabular}
\end{adjustbox}
\label{sst-prompt-types}
\end{table}

\section{Conclusion and Future Work}

This study demonstrates that carefully designed prompts enable VLMs to extract structured information from BPMN diagram images. We conducted a systematic evaluation across four complementary settings, entities, relations, types, and name-only, alongside a range of prompting strategies, showing how both the evaluation perspective and prompt design shape performance. Our results further reveal that OCR integration provides selective gains, particularly for weaker models that struggle to capture structured semantics. To support nuanced comparisons, we also introduced a tiered evaluation framework (strict and relaxed variants), which highlights robustness and element-level accuracy.

For future work, we plan to investigate fine-tuning on larger and more diverse BPMN datasets, with a focus on cross-diagram generalization. We will also explore advanced prompt engineering and multi-step reasoning techniques to improve semantic parsing and ensure consistent conversion of BPMN images into executable XML, thereby enabling automated process mining and seamless integration into real-world business environments.


\begin{acks}
We thank the maintainers of Hugging Face, OpenAI, Mistral, and other VLM developers for access to their models and APIs. The authors acknowledge the use of generative AI tools (e.g., ChatGPT) for support in code implementation and text editing; all conceptual contributions and experimental results are original and have been independently verified. This research is supported by the Advanced Research and Engineering Centre (ARC) in Northern Ireland, funded by PwC and Invest NI. The views expressed are those of the authors and do not necessarily represent those of ARC or the funding organisations. 
\end{acks}

\printbibliography

@String{Computing = "Computing" }

@String{Computer = "{IEEE} Computer" }

@String{Springer = "Springer-Verlag" }

@ArtifactSoftware{R,
    title = {R: A Language and Environment for Statistical Computing},
    author = {{R Core Team}},
    organization = {R Foundation for Statistical Computing},
    address = {Vienna, Austria},
    year = {2019},
    url = {https://www.R-project.org/},
}

@misc{bpmn-for-research,
  author       = {Camunda Services GmbH},
  title        = {BPMN for Research},
  howpublished = {\url{https://github.com/camunda/bpmn-for-research}},
  year         = {2015},
  note         = {Accessed: 2025-04-04}
}

@misc{bpmn-js,
  author       = {Camunda Services GmbH},
  title        = {bpmn-js: A BPMN 2.0 Rendering Toolkit and Web Modeler},
  howpublished = {\url{https://github.com/bpmn-io/bpmn-js}},
  year         = {2014},
  note         = {Accessed: 2025-04-04}
}

@misc{qwen2.5-VL,
    title = {Qwen2.5-VL},
    url = {https://qwenlm.github.io/blog/qwen2.5-vl/},
    author = {Qwen Team},
    month = {January},
    year = {2025}
}

@article{Qwen-VL,
  title={Qwen-VL: A Versatile Vision-Language Model for Understanding, Localization, Text Reading, and Beyond},
  author={Bai, Jinze and Bai, Shuai and Yang, Shusheng and Wang, Shijie and Tan, Sinan and Wang, Peng and Lin, Junyang and Zhou, Chang and Zhou, Jingren},
  journal={arXiv preprint arXiv:2308.12966},
  year={2023}
}

@article{agrawal2024pixtral,
  title={Pixtral 12B},
  author={Agrawal, Pravesh and Antoniak, Szymon and Hanna, Emma Bou and Bout, Baptiste and Chaplot, Devendra and Chudnovsky, Jessica and Costa, Diogo and De Monicault, Baudouin and Garg, Saurabh and Gervet, Theophile and others},
  journal={arXiv preprint arXiv:2410.07073},
  year={2024}
}

@article{team2025gemma,
  title={Gemma 3 Technical Report},
  author={Team, Gemma and Kamath, Aishwarya and Ferret, Johan and Pathak, Shreya and Vieillard, Nino and Merhej, Ramona and Perrin, Sarah and Matejovicova, Tatiana and Ram{\'e}, Alexandre and Rivi{\`e}re, Morgane and others},
  journal={arXiv preprint arXiv:2503.19786},
  year={2025}
}

@article{hurst2024gpt,
  title={Gpt-4o system card},
  author={Hurst, Aaron and Lerer, Adam and Goucher, Adam P and Perelman, Adam and Ramesh, Aditya and Clark, Aidan and Ostrow, AJ and Welihinda, Akila and Hayes, Alan and Radford, Alec and others},
  journal={arXiv preprint arXiv:2410.21276},
  year={2024}
}

@misc{meta2024llama3.2-vision,
  title={Llama 3.2 11B Vision Instruct Model},
  author={Meta AI},
  year={2024},
  url={https://huggingface.co/meta-llama/Llama-3.2-11B-Vision-Instruct}
}

@misc{bpmn-standard,
  title = {Business Process Model and Notation (BPMN) Specification},
  author = {Object Management Group},
  year = {2014},
  url = {https://www.omg.org/spec/BPMN/2.0}
}

@book{laguna2018business,
  title={Business process modeling, simulation and design},
  author={Laguna, Manuel and Marklund, Johan},
  year={2018},
  publisher={Chapman and Hall/CRC}
}

@article{matos2009migrating,
  title={Migrating legacy systems to service-oriented architectures},
  author={Matos, Carlos and Heckel, Reiko},
  journal={Electronic Communications of the EASST},
  volume={16},
  year={2009}
}

@article{van2012process,
  title={Process mining: Overview and opportunities},
  author={Van Der Aalst, Wil},
  journal={ACM Transactions on Management Information Systems (TMIS)},
  volume={3},
  number={2},
  pages={1--17},
  year={2012},
  publisher={ACM New York, NY, USA}
}

@book{allweyer2016bpmn,
  title={BPMN 2.0: introduction to the standard for business process modeling},
  author={Allweyer, Thomas},
  year={2016},
  publisher={BoD--Books on Demand}
}

@article{kocbek2015business,
  title={Business process model and notation: The current state of affairs},
  author={Kocbek, Mateja and Jo{\v{s}}t, Gregor and Heri{\v{c}}ko, Marjan and Polan{\v{c}}i{\v{c}}, Gregor},
  journal={Computer Science and Information Systems},
  volume={12},
  number={2},
  pages={509--539},
  year={2015}
}

@article{ko2009business,
  title={Business process management (BPM) standards: a survey},
  author={Ko, Ryan KL and Lee, Stephen SG and Wah Lee, Eng},
  journal={Business process management journal},
  volume={15},
  number={5},
  pages={744--791},
  year={2009},
  publisher={Emerald Group Publishing Limited}
}

@inproceedings{minor2024retrieval,
  title={Retrieval augmented generation with LLMs for explaining business process models},
  author={Minor, Mirjam and Kaucher, Eduard},
  booktitle={International Conference on Case-Based Reasoning},
  pages={175--190},
  year={2024},
  organization={Springer}
}

@inproceedings{bellan2022extracting,
  title={Extracting business process entities and relations from text using pre-trained language models and in-context learning},
  author={Bellan, Patrizio and Dragoni, Mauro and Ghidini, Chiara},
  booktitle={International Conference on Enterprise Design, Operations, and Computing},
  pages={182--199},
  year={2022},
  organization={Springer}
}

@inproceedings{friedrich2011process,
  title={Process model generation from natural language text},
  author={Friedrich, Fabian and Mendling, Jan and Puhlmann, Frank},
  booktitle={Advanced Information Systems Engineering: 23rd International Conference, CAiSE 2011, London, UK, June 20-24, 2011. Proceedings 23},
  pages={482--496},
  year={2011},
  organization={Springer}
}

@inproceedings{grohs2023large,
  title={Large language models can accomplish business process management tasks},
  author={Grohs, Michael and Abb, Luka and Elsayed, Nourhan and Rehse, Jana-Rebecca},
  booktitle={International Conference on Business Process Management},
  pages={453--465},
  year={2023},
  organization={Springer}
}

@inproceedings{kopke2024efficient,
  title={Efficient llm-based conversational process modeling},
  author={K{\"o}pke, Julius and Safan, Aya},
  booktitle={International Conference on Business Process Management},
  pages={259--270},
  year={2024},
  organization={Springer}
}

@article{corradini2024technique,
  title={A technique for discovering BPMN collaboration diagrams},
  author={Corradini, Flavio and Pettinari, Sara and Re, Barbara and Rossi, Lorenzo and Tiezzi, Francesco},
  journal={Software and Systems Modeling},
  pages={1--21},
  year={2024},
  publisher={Springer}
}

@inproceedings{nour2024nala2bpmn,
  title={Nala2BPMN: Automating BPMN Model Generation with Large Language Models},
  author={Nour Eldin, Ali and Assy, Nour and Anesini, Olan and Dalmas, Benjamin and Gaaloul, Walid},
  booktitle={International Conference on Cooperative Information Systems},
  pages={398--404},
  year={2024},
  organization={Springer}
}

@article{licardo2024method,
  title={A Method for Extracting BPMN Models from Textual Descriptions Using Natural Language Processing},
  author={Licardo, Josip Tomo and Tankovi{\'c}, Nikola and Etinger, Darko},
  journal={Procedia computer science},
  volume={239},
  pages={483--490},
  year={2024},
  publisher={Elsevier}
}

@article{sholiq2022generating,
  title={Generating BPMN diagram from textual requirements},
  author={Sholiq, Sholiq and Sarno, Riyanarto and Astuti, Endang Siti},
  journal={Journal of King Saud University-Computer and Information Sciences},
  volume={34},
  number={10},
  pages={10079--10093},
  year={2022},
  publisher={Elsevier}
}

@article{fu2025vita,
  title={Vita-1.5: Towards gpt-4o level real-time vision and speech interaction},
  author={Fu, Chaoyou and Lin, Haojia and Wang, Xiong and Zhang, Yi-Fan and Shen, Yunhang and Liu, Xiaoyu and Cao, Haoyu and Long, Zuwei and Gao, Heting and Li, Ke and others},
  journal={arXiv preprint arXiv:2501.01957},
  year={2025}
}

@inproceedings{2023GPT4VisionSC,
  title={GPT-4V(ision) System Card},
  author={},
  year={2023},
  url={https://api.semanticscholar.org/CorpusID:263218031}
}

@article{lu2023mathvista,
  title={Mathvista: Evaluating mathematical reasoning of foundation models in visual contexts},
  author={Lu, Pan and Bansal, Hritik and Xia, Tony and Liu, Jiacheng and Li, Chunyuan and Hajishirzi, Hannaneh and Cheng, Hao and Chang, Kai-Wei and Galley, Michel and Gao, Jianfeng},
  journal={arXiv preprint arXiv:2310.02255},
  year={2023}
}

@inproceedings{zhang2024mathverse,
  title={Mathverse: Does your multi-modal llm truly see the diagrams in visual math problems?},
  author={Zhang, Renrui and Jiang, Dongzhi and Zhang, Yichi and Lin, Haokun and Guo, Ziyu and Qiu, Pengshuo and Zhou, Aojun and Lu, Pan and Chang, Kai-Wei and Qiao, Yu and others},
  booktitle={European Conference on Computer Vision},
  pages={169--186},
  year={2024},
  organization={Springer}
}

@article{chen2024far,
  title={How far are we to gpt-4v? closing the gap to commercial multimodal models with open-source suites},
  author={Chen, Zhe and Wang, Weiyun and Tian, Hao and Ye, Shenglong and Gao, Zhangwei and Cui, Erfei and Tong, Wenwen and Hu, Kongzhi and Luo, Jiapeng and Ma, Zheng and others},
  journal={Science China Information Sciences},
  volume={67},
  number={12},
  pages={220101},
  year={2024},
  publisher={Springer}
}

@article{roberts2024image2struct,
  title={Image2struct: Benchmarking structure extraction for vision-language models},
  author={Roberts, Josselin and Lee, Tony and Wong, Chi Heem and Yasunaga, Michihiro and Mai, Yifan and Liang, Percy S},
  journal={Advances in Neural Information Processing Systems},
  volume={37},
  pages={115058--115097},
  year={2024}
}

@inproceedings{lee2023pix2struct,
  title={Pix2struct: Screenshot parsing as pretraining for visual language understanding},
  author={Lee, Kenton and Joshi, Mandar and Turc, Iulia Raluca and Hu, Hexiang and Liu, Fangyu and Eisenschlos, Julian Martin and Khandelwal, Urvashi and Shaw, Peter and Chang, Ming-Wei and Toutanova, Kristina},
  booktitle={International Conference on Machine Learning},
  pages={18893--18912},
  year={2023},
  organization={PMLR}
}

@misc{RapidOCR2021,
    title={{Rapid OCR}: OCR Toolbox},
    author={RapidAI Team},
    howpublished = {\url{https://github.com/RapidAI/RapidOCR}},
    year={2021}
}

@inproceedings{smith2007overview,
  title={An overview of the Tesseract OCR engine},
  author={Smith, Ray},
  booktitle={Ninth international conference on document analysis and recognition (ICDAR 2007)},
  volume={2},
  pages={629--633},
  year={2007},
  organization={IEEE}
}

@inproceedings{wolf2020transformers,
  title={Transformers: State-of-the-art natural language processing},
  author={Wolf, Thomas and Debut, Lysandre and Sanh, Victor and Chaumond, Julien and Delangue, Clement and Moi, Anthony and Cistac, Pierric and Rault, Tim and Louf, R{\'e}mi and Funtowicz, Morgan and others},
  booktitle={Proceedings of the 2020 conference on empirical methods in natural language processing: system demonstrations},
  pages={38--45},
  year={2020}
}

@inproceedings{antinori2022bpmn,
  title={BPMN-Redrawer: From Images to BPMN Models.},
  author={Antinori, Alessandro and Coltrinari, Riccardo and Corradini, Flavio and Fornari, Fabrizio and Re, Barbara and Scarpetta, Marco and others},
  year={2022}
}

@inproceedings{devlin2019bert,
  title={Bert: Pre-training of deep bidirectional transformers for language understanding},
  author={Devlin, Jacob and Chang, Ming-Wei and Lee, Kenton and Toutanova, Kristina},
  booktitle={Proceedings of the 2019 conference of the North American chapter of the association for computational linguistics: human language technologies, volume 1 (long and short papers)},
  pages={4171--4186},
  year={2019}
}

@article{kourani2024promoai,
  title={ProMoAI: process modeling with generative AI},
  author={Kourani, Humam and Berti, Alessandro and Schuster, Daniel and Van der Aalst, Wil MP},
  journal={arXiv preprint arXiv:2403.04327},
  year={2024}
}

@article{lin2024mao,
  title={MAO: A Framework for Process Model Generation with Multi-Agent Orchestration},
  author={Lin, Leilei and Jin, Yumeng and Zhou, Yingming and Chen, Wenlong and Qian, Chen},
  journal={arXiv preprint arXiv:2408.01916},
  year={2024}
}

@article{fagnoni2024opus,
  title={Opus: A Large Work Model for Complex Workflow Generation},
  author={Fagnoni, Th{\'e}o and Mesbah, Bellinda and Altin, Mahsun and Kingston, Phillip},
  journal={arXiv preprint arXiv:2412.00573},
  year={2024}
}

@article{medhi2024target,
  title={Target Prompting for Information Extraction with Vision Language Model},
  author={Medhi, Dipankar},
  journal={arXiv preprint arXiv:2408.03834},
  year={2024}
}

@article{khan2024fine,
  title={Fine-Tuning Vision-Language Model for Automated Engineering Drawing Information Extraction},
  author={Khan, Muhammad Tayyab and Chen, Lequn and Ng, Ye Han and Feng, Wenhe and Tan, Nicholas Yew Jin and Moon, Seung Ki},
  journal={arXiv preprint arXiv:2411.03707},
  year={2024}
}

@inproceedings{scius2025zero,
  title={Zero-Shot Prompting and Few-Shot Fine-Tuning: Revisiting Document Image Classification Using Large Language Models},
  author={Scius-Bertrand, Anna and Jungo, Michael and V{\"o}gtlin, Lars and Spat, Jean-Marc and Fischer, Andreas},
  booktitle={International Conference on Pattern Recognition},
  pages={152--166},
  year={2025},
  organization={Springer}
}

@article{masry2022chartqa,
  title={Chartqa: A benchmark for question answering about charts with visual and logical reasoning},
  author={Masry, Ahmed and Long, Do Xuan and Tan, Jia Qing and Joty, Shafiq and Hoque, Enamul},
  journal={arXiv preprint arXiv:2203.10244},
  year={2022}
}

@article{tian2024chartgpt,
  title={Chartgpt: Leveraging llms to generate charts from abstract natural language},
  author={Tian, Yuan and Cui, Weiwei and Deng, Dazhen and Yi, Xinjing and Yang, Yurun and Zhang, Haidong and Wu, Yingcai},
  journal={IEEE Transactions on Visualization and Computer Graphics},
  year={2024},
  publisher={IEEE}
}

@inproceedings{tannert2023flowchartqa,
  title={FlowchartQA: the first large-scale benchmark for reasoning over flowcharts},
  author={Tannert, Simon and Feighelstein, Marcelo G and Bogojeska, Jasmina and Shtok, Joseph and Arbelle, Assaf and Staar, Peter WJ and Schumann, Anika and Kuhn, Jonas and Karlinsky, Leonid},
  booktitle={Proceedings of the 1st Workshop on Linguistic Insights from and for Multimodal Language Processing},
  pages={34--46},
  year={2023}
}

@article{xu2023chartbench,
  title={Chartbench: A benchmark for complex visual reasoning in charts},
  author={Xu, Zhengzhuo and Du, Sinan and Qi, Yiyan and Xu, Chengjin and Yuan, Chun and Guo, Jian},
  journal={arXiv preprint arXiv:2312.15915},
  year={2023}
}

@inproceedings{hu2024novachart,
  title={NovaChart: A Large-scale Dataset towards Chart Understanding and Generation of Multimodal Large Language Models},
  author={Hu, Linmei and Wang, Duokang and Pan, Yiming and Yu, Jifan and Shao, Yingxia and Feng, Chong and Nie, Liqiang},
  booktitle={Proceedings of the 32nd ACM International Conference on Multimedia},
  pages={3917--3925},
  year={2024}
}

@article{masry2024chartinstruct,
  title={Chartinstruct: Instruction tuning for chart comprehension and reasoning},
  author={Masry, Ahmed and Shahmohammadi, Mehrad and Parvez, Md Rizwan and Hoque, Enamul and Joty, Shafiq},
  journal={arXiv preprint arXiv:2403.09028},
  year={2024}
}

@article{han2023chartllama,
  title={Chartllama: A multimodal llm for chart understanding and generation},
  author={Han, Yucheng and Zhang, Chi and Chen, Xin and Yang, Xu and Wang, Zhibin and Yu, Gang and Fu, Bin and Zhang, Hanwang},
  journal={arXiv preprint arXiv:2311.16483},
  year={2023}
}

@article{masry2024chartgemma,
  title={Chartgemma: Visual instruction-tuning for chart reasoning in the wild},
  author={Masry, Ahmed and Thakkar, Megh and Bajaj, Aayush and Kartha, Aaryaman and Hoque, Enamul and Joty, Shafiq},
  journal={arXiv preprint arXiv:2407.04172},
  year={2024}
}

@article{xia2024chartx,
  title={Chartx \& chartvlm: A versatile benchmark and foundation model for complicated chart reasoning},
  author={Xia, Renqiu and Zhang, Bo and Ye, Hancheng and Yan, Xiangchao and Liu, Qi and Zhou, Hongbin and Chen, Zijun and Ye, Peng and Dou, Min and Shi, Botian and others},
  journal={arXiv preprint arXiv:2402.12185},
  year={2024}
}

@article{singh2024flowvqa,
  title={Flowvqa: Mapping multimodal logic in visual question answering with flowcharts},
  author={Singh, Shubhankar and Chaurasia, Purvi and Varun, Yerram and Pandya, Pranshu and Gupta, Vatsal and Gupta, Vivek and Roth, Dan},
  journal={arXiv preprint arXiv:2406.19237},
  year={2024}
}

@article{wilcoxon1945individual,
  title={Individual comparisons by ranking methods},
  author={Wilcoxon, Frank},
  journal={Biometrics bulletin},
  volume={1},
  number={6},
  pages={80--83},
  year={1945},
  publisher={JSTOR}
}

@article{friedman1937use,
  title={The use of ranks to avoid the assumption of normality implicit in the analysis of variance},
  author={Friedman, Milton},
  journal={Journal of the american statistical association},
  volume={32},
  number={200},
  pages={675--701},
  year={1937},
  publisher={Taylor \& Francis}
}

@article{kendall1939problem,
  title={The problem of m rankings},
  author={Kendall, Maurice G and Smith, B Babington},
  journal={The annals of mathematical statistics},
  volume={10},
  number={3},
  pages={275--287},
  year={1939},
  publisher={JSTOR}
}

@book{cohen2013statistical,
  title={Statistical power analysis for the behavioral sciences},
  author={Cohen, Jacob},
  year={2013},
  publisher={routledge}
}

@article{schafer2022sketch2process,
  title={Sketch2process: End-to-end BPMN sketch recognition based on neural networks},
  author={Sch{\"a}fer, Bernhard and Van Der Aa, Han and Leopold, Henrik and Stuckenschmidt, Heiner},
  journal={IEEE Transactions on Software Engineering},
  volume={49},
  number={4},
  pages={2621--2641},
  year={2022},
  publisher={IEEE}
}

\end{document}